\documentclass[11pt]{article}
\usepackage[T1]{fontenc}
\usepackage{lmodern}
\usepackage[margin=1in]{geometry}
\usepackage{setspace}
\RequirePackage{amsthm,amsmath,amsfonts,amssymb}
\RequirePackage{mathrsfs,mathtools,bm,booktabs,graphicx,float,placeins,enumitem}
\RequirePackage{algorithm,algpseudocode}
\RequirePackage[authoryear,round]{natbib}
\RequirePackage[hidelinks]{hyperref}
\RequirePackage{xcolor}
\numberwithin{equation}{section}
\theoremstyle{plain}
\newtheorem{theorem}{Theorem}[section]
\newtheorem{lemma}{Lemma}[section]
\newtheorem{proposition}{Proposition}[section]

\theoremstyle{definition}
\newtheorem{assumption}{Assumption}[section]

\newcommand{\M}{\mathcal M}
\newcommand{\R}{\mathbb R}
\newcommand{\E}{\mathbb E}
\newcommand{\Pp}{\mathbb P}
\newcommand{\Vol}{\operatorname{Vol}}
\newcommand{\Cov}{\operatorname{Cov}}

\newcommand{\dd}{\,\mathrm d}
\newcommand{\1}{\mathbf 1}
\newcommand{\norm}[1]{\left\lVert #1\right\rVert}
\newcommand{\inner}[2]{\left\langle #1,#2\right\rangle}
\newcommand{\Tau}{\mathcal P}
\newcommand{\wh}{\widehat}

\title{Riemannian Simultaneous Inference for Tangent Vector Field Regression}
\author{Xiaotian Chang\thanks{Division of Mathematical Sciences, Nanyang Technological University.}
\quad Yangdi Jiang\footnotemark[1]
\quad Qirui Hu\thanks{\begin{tabular}[t]{@{}l@{}}
School of Statistics and Data Science, Shanghai University of Finance and Economics.\\
Department of Mathematics, Ruhr-Universit\"at Bochum\\
Corresponding author. Email: \href{mailto:huqirui@mail.shufe.edu.cn}{huqirui@mail.shufe.edu.cn}
\end{tabular}}}
\date{}
\begin{document}
\maketitle
\begin{abstract}
We consider nonparametric tangent vector field regression on a Riemannian manifold without boundary.
Because responses at different
points lie in different tangent spaces, the proposed kernel estimator first
parallel transports nearby responses to the target tangent space and then
forms a volume-corrected local average. We first derive its uniform second-order
bias, finite-bandwidth covariance, and stochastic rate.  For simultaneous
inference, the tangent norm is written as a supremum over the unit tangent
bundle.  Exact covariance whitening gives a unit-variance Gaussian
field whose correlation length is of order $h$ along the base manifold and of
order one along the fibre.  Its local covariance geometry leads to a
Gumbel limit with an explicit intrinsic constant.  Combining this limit with
Gaussian approximation and cross-fitted covariance estimation yields a
feasible simultaneous confidence tube for the regression field. We further discuss improved finite-sample inference with bandwidth selection and high-order bias corrections.
Simulations on various manifolds support the proposed inference procedure.  A randomized reconstruction of global wind data
illustrates how the tube's cross-sections describe spatially varying uncertainty.

\end{abstract}


\section{Introduction}\label{SEC:Introduction}
Many scientific measurements describe local motion on a curved domain.
Their locations lie on a Riemannian manifold $\M$, while their values lie in the tangent
spaces at those locations.  Horizontal wind provides a familiar example:
a geographic location is a point $x\in S^2$, and the wind velocity is an
element of the tangent plane $T_xS^2$.  Estimating the mean circulation and
quantifying uncertainty over the globe therefore requires statistical methods
for an entire tangent vector field, rather than separate analyses of vectors
in one fixed Euclidean space
\citep{fanpaulleematsuo2018,robertnicoudkrauseborovitskiy2024}.

\begin{figure}[t]
\centering
\begin{minipage}[t]{0.42\textwidth}
\vspace{0pt}\centering
\includegraphics[width=\linewidth]{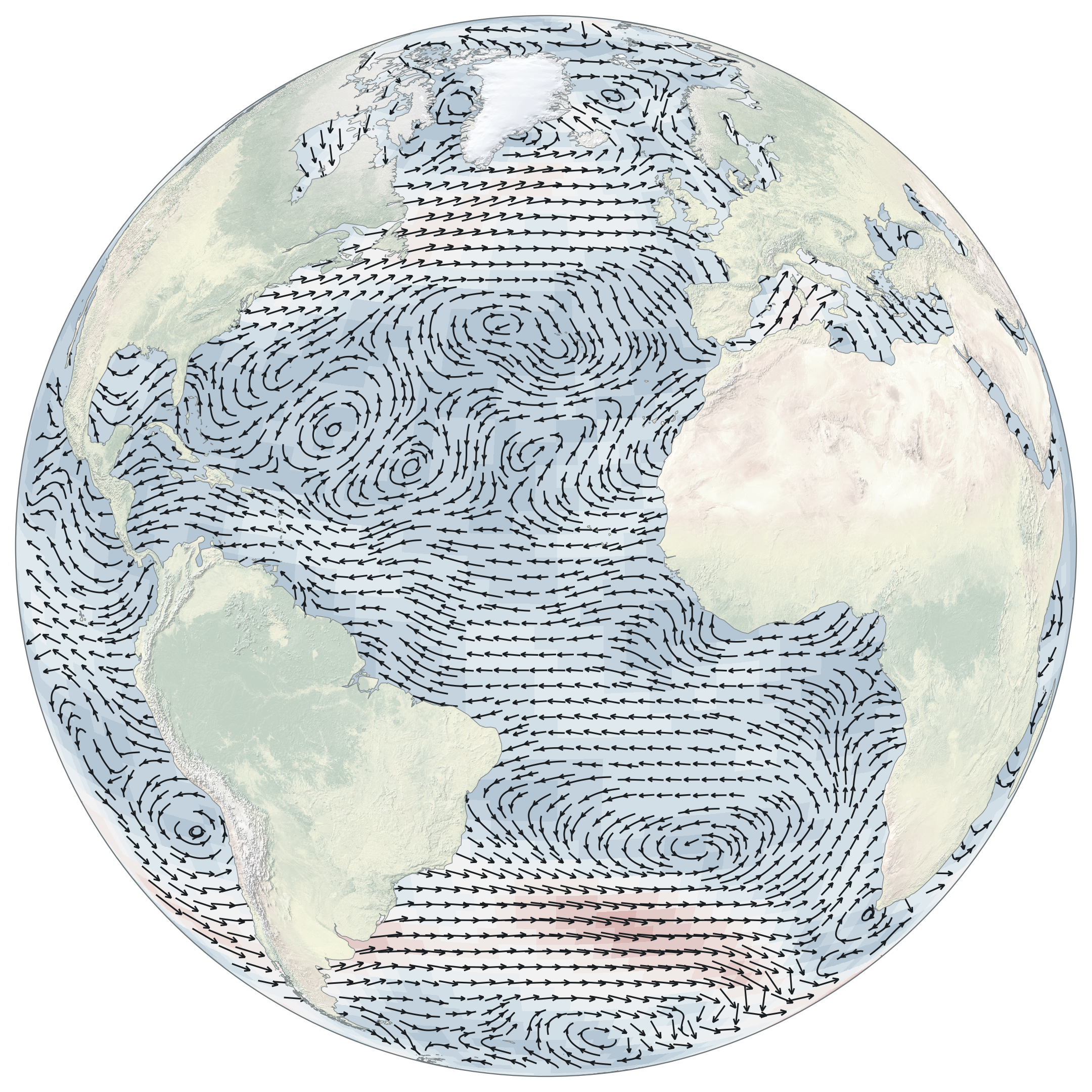}\par
\vspace{-3pt}
\includegraphics[width=0.92\linewidth]{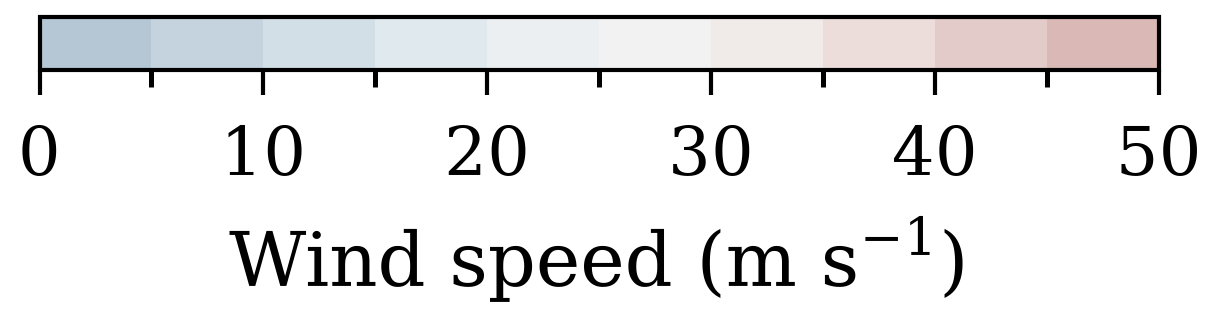}
\end{minipage}\hfill
\begin{minipage}[t]{0.56\textwidth}
\vspace{0pt}\centering
\includegraphics[width=\linewidth]{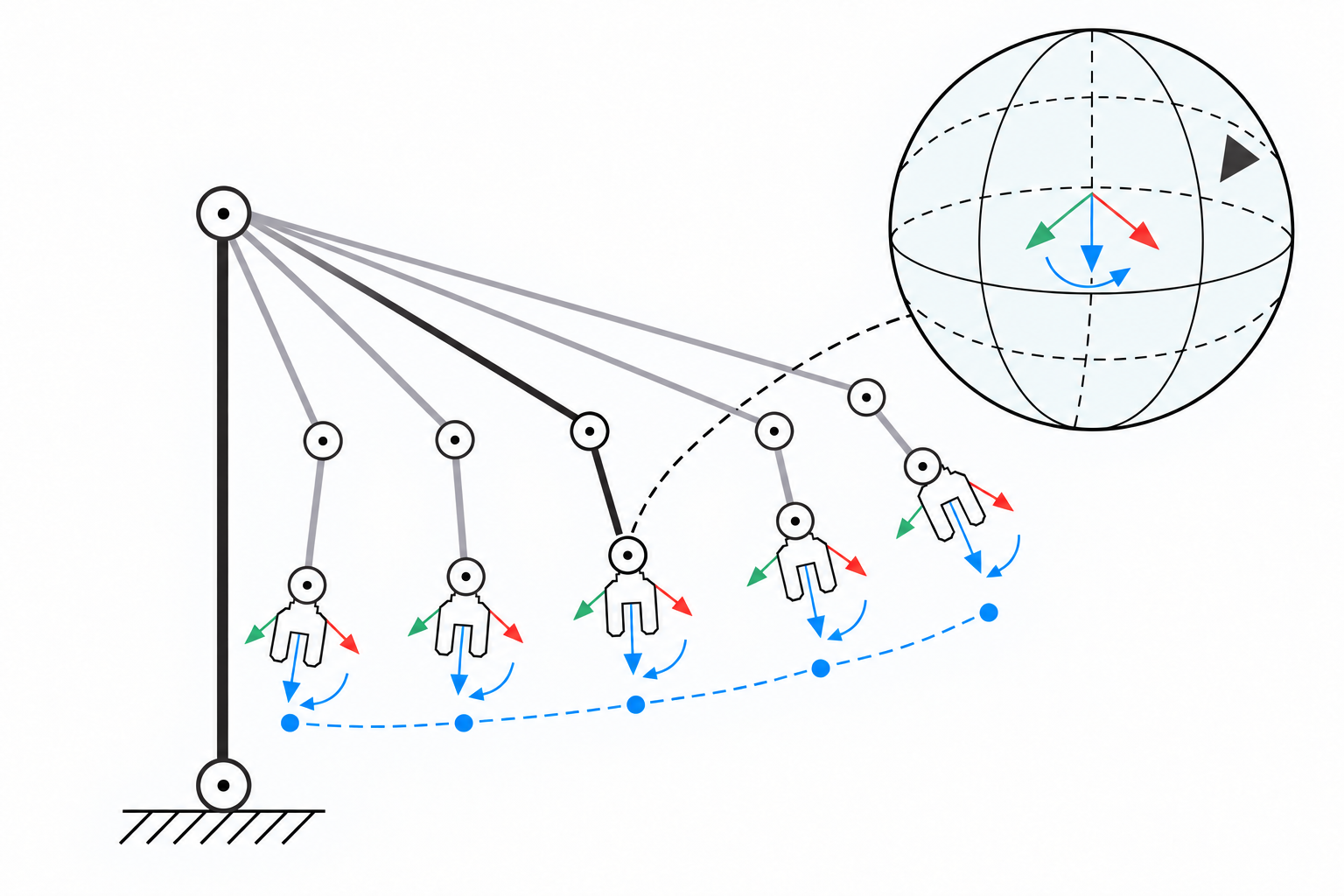}\par
\smallskip
{\footnotesize
Orientation $R\in SO\left(3\right)$; body angular velocity $\omega\in\R^3$.\par
Tangent response: $Y=R\left[\omega\right]_\times\in T_RSO\left(3\right)$.\par
Target: mean tangent velocity $V\left(R\right)=\E\left(Y\mid R\right)$,\par
with simultaneous uncertainty.\par}
\end{minipage}
\caption{Tangent-vector observations on a sphere and a rotation group.
Left: 500-hPa wind on 1 July 2024, 12:00 UTC;
color indicates speed and arrows indicate direction.
Right: schematic end-effector orientations and angular velocities on $SO\left(3\right)$.}
\label{FIG:IntroApplications}
\end{figure}

The same data structure occurs in robotics.  The full orientation of an end
effector is a rotation $R\in SO\left(3\right)$, and a body angular velocity
$\omega\in\R^3$ determines the tangent velocity
$R\left[\omega\right]_\times\in T_RSO\left(3\right)$, where $\left[\omega\right]_\times v=\omega\times v$.
Orientation-dependent dynamical systems are used to describe end-effector
motion \citep{beikmohammadi2024contractive}.
Repeated orientation--velocity observations can therefore be used to estimate
mean rotational motion and its uncertainty as orientation varies.
Figure~\ref{FIG:IntroApplications} illustrates these two motivating settings\footnote{The wind image uses NCEP--NCAR Reanalysis~1 on a $2.5^\circ$
grid, provided by NOAA PSL, Boulder, Colorado, USA (\url{https://psl.noaa.gov});
land imagery is from Natural Earth.  This snapshot differs from the data
in Section~\ref{SEC:NCEPWind}.};
the robot panel is schematic. We further present the data analysis concerning
planetary-scale wind reconstruction in Section \ref{SEC:NCEPWind}.

To place these applications in a common statistical framework, we observe
independent copies of a pair $\left(X,Y\right)$ satisfying
\begin{equation}\label{EQ:Model}
  X_i\in\M,
  \qquad
  Y_i\in T_{X_i}\M,
  \qquad
  Y_i=V\left(X_i\right)+\eta_i,
  \qquad
  \E\left(\eta_i\mid X_i\right)=0,
\end{equation}
where $X$ has density $p$ with respect to Riemannian volume, and the unknown
regression field is $V\left(x\right)=\E\left(Y\mid X=x\right)\in T_x\M$.  Thus $V$ is a section
of the tangent bundle: it assigns a vector to each location in its own tangent
space.  The conditional error covariance may vary with location.

Our first objective is to estimate $V$ uniformly over $\M$ using a kernel estimator.  Our principal objective is simultaneous inference: given
$\alpha\in\left(0,1\right)$, construct regions $\mathcal C_{n,1-\alpha}\left(x\right)\subset T_x\M$
such that
\[
 \Pp\left\{V\left(x\right)\in\mathcal C_{n,1-\alpha}\left(x\right)
       \text{ for every }x\in\M\right\}\longrightarrow1-\alpha.
\]
These regions form a confidence tube around the estimated field, covering
all locations and tangent directions simultaneously.  We allow a
nonparametric mean field and spatially varying error covariance on a known
compact manifold.

Kernel methods on manifolds provide a natural starting point.  Volume-corrected
kernel density estimation and scalar-response regression on closed manifolds
were studied by \citet{pelletier2005density,pelletier2006regression};
\citet{henryrodriguez2009} developed robust regression in the same geometric
setting.  Tangent responses introduce a further issue: both the location and
the space containing the response vary.  \citet{singerwu2012vector} use local
alignments to construct vector diffusion maps and relate their limit to the
connection Laplacian, with applications to vector-field interpolation and
regression.  Their manifold-learning setting differs from the known-manifold
random-design model considered here.  \citet{dassnasel2026} study concentration
for bundle-valued statistics transported to a reference fibre.  Such
concentration results address a different objective from calibrating a
shrinking-bandwidth regression maximum over all target fibres.

A complementary literature models tangent fields through Gaussian processes.
On the sphere, \citet{fanpaulleematsuo2018} construct covariance models from
surface gradients and curls of scalar potentials, with likelihood-based
parameter estimation.
\citet{hutchinson2021vector} develop gauge-independent projected kernels,
whereas \citet{robertnicoudkrauseborovitskiy2024} construct intrinsic
Hodge--Mat\'ern Gaussian vector fields.  The latter two constructions
provide Gaussian process priors for Bayesian prediction and uncertainty
quantification: the unknown field is modeled as random, and inference uses
its conditional distribution given the observations.  Our objective is instead
frequentist simultaneous coverage for a fixed unknown regression field under
independent noisy observations; the Gaussian field below is a reference law
for estimation error, not a prior on $V$.

Simultaneous inference has a long history in Euclidean statistics.
\citet{bickelrosenblatt1973global} study global deviations of density
estimators, and \citet{johnston1982maximal} obtain maximal-deviation limits for
nonparametric regression estimators.  \citet{hardlemarron1991bootstrap}
construct simultaneous regression error bars by residual bootstrap, providing
an alternative to analytic extreme-value calibration.  Modern Gaussian
approximation and anti-concentration methods support confidence-band
construction without requiring an extreme-value limit
\citep{chernozhukov2014bands,chernozhukov2015comparison,chernozhuokov2022improved}.
For manifold-indexed Gaussian fields, \citet{qiaopolonik2018} study extrema
under rescaling, and \citet{qiao2021extremes} develops excursion and extreme-value
results for locally stationary Gaussian and chi fields.  Related statistical
uses include confidence regions for density ridges \citep{qiao2021ridges}
and multivariate random-field inference \citep{taylorworsley2008}.

Two difficulties remain in the present problem.  First, nearby responses
must be brought into a common tangent space before averaging.  A global
orthonormal frame need not exist, and coordinatewise confidence bands would
depend on the frames chosen.  Curvature and a nonuniform design also affect
the local mean and covariance.  Second, simultaneous inference requires
the dependence between estimation errors at different locations.  Gaussian
approximations defined separately in local frames must agree where those
frames overlap.  Their excursion probabilities must then be combined to
obtain one threshold for all locations and tangent directions.

\label{SEC:Contributions}
We address the first difficulty by transporting nearby responses along
short geodesics before taking a volume-corrected kernel average.  Its
second-order bias identifies the effects of the covariant derivatives of
$V$ and the design density.  Uniform stochastic bounds then separate the
local noise, transported-signal variation and denominator error.

For simultaneous inference, we construct one Gaussian reference section
whose covariance matches the leading estimation error at every pair of
locations.  A shared Hilbert-space representation ensures that its local
representations agree on chart overlaps.  Expressing its norm as a maximum
over unit tangent directions allows us to compare its maximum with the
regression maximum on a common finite grid.  A local covariance expansion
then determines high-excursion probabilities, which combine into a global
Gumbel law with an explicit constant.  This gives an analytic critical
value for the confidence tube.

To make the confidence tube feasible, we estimate the error covariance from cross-fitted
residuals at each observation location.  These covariance estimates
are transported to the target tangent space and combined with squared
kernel weights, as required for the covariance of a weighted mean.
Separate bandwidths for regression, residual prediction and covariance
smoothing make the covariance estimate sufficiently accurate for this
calibration.  Undersmoothing gives coverage
for $V$; a separate target-local construction gives inference for the smoothed
field when spatial resolution is part of the estimand.  Simulations and the
wind reconstruction illustrate the resulting confidence tubes.

The rest of the paper is organized as follows. Section~\ref{SEC:Setup} introduces the model, geometry, estimator and assumptions.
Section~\ref{SEC:Estimation} establishes the estimation results, and
Section~\ref{SEC:Inference} develops simultaneous inference.
Sections~\ref{SEC:SimulationDesign} and~\ref{SEC:NCEPWind} present the simulations
and wind reconstruction.  The supplementary material follows
the main results in proof order and provides implementation details.

\section{Methodology}\label{SEC:Setup}

We first specify the geometry needed to compare responses at different
locations and then define the kernel estimator.  The geometry is assumed known.

Let $\left(\M,g\right)$ be a compact, connected, smooth Riemannian manifold of dimension
$d\ge2$ without boundary.  Write $\rho$ for geodesic distance, $\Vol$ for
Riemannian volume, $\inner{\cdot}{\cdot}_x$ and $\norm{\cdot}_x$ for the
inner product and norm on $T_x\M$, and $\iota_\M>0$ for the injectivity
radius.  The exponential map $\exp_x$ sends an initial tangent
vector $u$ to the time-one endpoint of the geodesic starting at $x$ with
velocity $u$.  On a tangent ball of radius less than $\iota_\M$, it is a
diffeomorphism onto a normal neighborhood of $x$.  Its inverse is the
logarithm map $\log_x$.  If $\rho\left(x,q\right)<\iota_\M$, let
$\Tau_{q\to x}$ denote Levi--Civita parallel transport along
the unique minimizing geodesic.  Transport is an isometry and
$\Tau_{q\to x}^{-1}=\Tau_{x\to q}$.

For $q=\exp_x\left(u\right)$ in a normal neighborhood, define the volume density by
\[
 \dd\Vol\left(q\right)=\Theta_x\left(q\right)\,\dd u,
\]
where $\dd u$ is Lebesgue measure induced by the tangent inner product.
On uniformly small normal neighborhoods, $\Theta_x\left(q\right)$ is bounded above
and away from zero.  Its reciprocal removes volume distortion in the
kernel calculation.  We write $\nabla$ for the Levi--Civita covariant
derivative and $\Gamma\left(T\M\right)$ for the sections of the tangent bundle.
The endomorphism bundle $\operatorname{End}\left(T\M\right)$ has at $x$
the fibre of linear maps from $T_x\M$ to itself; a section assigns one such
map to every point.  The covariance field below is an example.
A local orthonormal frame supplies coordinates, but the section and its norm
do not depend on that frame.  For example,
$T_xS^2=\left\{u\in\R^3:x^\top u=0\right\}$ and $T_RSO\left(3\right)=\left\{RA:A^\top=-A\right\}$.

The unit tangent bundle, needed later for simultaneous inference, is
\[
 S\left(T\M\right)=\left\{\left(x,v\right):x\in\M,\ v\in T_x\M,\ \norm v_x=1\right\}.
\]
Its base has dimension $d$ and its fibres have dimension $d-1$.
Only local product coordinates are used; a global frame is not assumed.
See \citet{lee2018riemannian} for these geometric conventions.

Under model~\eqref{EQ:Model}, neither the density $p$ nor the regression
field $V$ is known.  The conditional error covariance is the self-adjoint
operator
\[
 \Sigma\left(x\right)=\E\left\{\eta_i\otimes\eta_i\mid X_i=x\right\}:T_x\M\to T_x\M,
 \qquad \left(u\otimes v\right)w=\inner{v}{w}_x u.
\]

To estimate at $x$, we first restrict attention to observations within
geodesic distance $h$ from $x$, transport their responses into $T_x\M$, and then
average them there.  The factor $\Theta_x\left(q\right)^{-1}$ removes the normal-coordinate
volume distortion from the local mean.  This is the geometric counterpart of
a Nadaraya--Watson average.

Let $K:\left[0,\infty\right)\to\R$ be radial and supported on $\left[0,1\right]$, and write
\begin{equation}\label{EQ:KernelConstants}
  K^\star\left(u\right)=K\left(\norm{u}\right),
  \qquad
  R\left(K\right)=\int_{\R^d}\left\{K^\star\left(u\right)\right\}^2\dd u,
  \qquad
  \mu_2\left(K\right)=\frac{1}{d}\int_{\R^d}\norm u^2K^\star\left(u\right)\dd u.
\end{equation}

For $0<h<\iota_\M$, define
\begin{equation*}
 W_h\left(x,q\right)=
 \begin{cases}
  \Theta_x\left(q\right)^{-1}K\left\{\rho\left(x,q\right)/h\right\},&\rho\left(x,q\right)<\iota_\M,\\
  0,&\rho\left(x,q\right)\ge\iota_\M.
 \end{cases}
\end{equation*}
Because $K$ is supported on $\left[0,1\right]$, $W_h\left(x,q\right)=0$ unless
$\rho\left(x,q\right)\le h$, where the short-geodesic transport is uniquely defined.

The empirical denominator and transported numerator are
\begin{align*}
  \wh p_h\left(x\right)
  &=\frac{1}{nh^d}\sum_{i=1}^nW_h\left(x,X_i\right),
  \\
  \wh N_h\left(x\right)
  &=\frac{1}{nh^d}\sum_{i=1}^nW_h\left(x,X_i\right)\Tau_{X_i\to x}Y_i
  \in T_x\M.
\end{align*}
Let $\underline p_n>0$ be a deterministic sequence with
$\underline p_n\downarrow0$.  For measurability on every sample, put
$\wh p_h^\dagger\left(x\right)=\wh p_h\left(x\right)\vee\underline p_n$ and define
\begin{equation}\label{EQ:Estimator}
  \wh V_h\left(x\right)=\frac{\wh N_h\left(x\right)}{\wh p_h^\dagger\left(x\right)}\in T_x\M.
\end{equation}
When the denominator floor is inactive, the normalized weights
$W_h\left(x,X_i\right)/\sum_jW_h\left(x,X_j\right)$ sum to one.
Under the assumptions below, the floor is uniformly inactive with
probability tending to one.

To analyze this local average, we introduce its population counterpart:
\begin{align*}
  p_h\left(x\right)&=\E\left\{\wh p_h\left(x\right)\right\},\\
  N_h\left(x\right)&=\frac{1}{h^d}\int_\M
  W_h\left(x,q\right)\Tau_{q\to x}V\left(q\right)p\left(q\right)\dd\Vol\left(q\right),\\
  V_h\left(x\right)&=\frac{N_h\left(x\right)}{p_h\left(x\right)}.
\end{align*}
\label{SEC:Assumptions}
The smoothed field $V_h$ is the population target of the local
average and separates stochastic error from smoothing bias.  The first
three assumptions below justify the local mean, covariance and Gaussian
calculations; the fourth makes $V_h-V$ negligible at the Gumbel scale for
simultaneous inference on $V$.  Conditions for the separate pilot and
covariance bandwidths appear in Section~\ref{SEC:CovarianceEstimation}. We first present and discuss some necessary assumptions for subsequent theoretical analysis.

 \begin{assumption}\label{ASS:Kernel}
The function $K^\star\left(u\right)=K\left(\norm{u}\right)$ is nonnegative, belongs to
$C_c^4\left(\R^d\right)$, is supported on the closed unit ball, and satisfies
$
  \int_{\R^d}K^\star\left(u\right)\dd u=1,
  0<R\left(K\right)<\infty.
$
\end{assumption}

 \begin{assumption}\label{ASS:Regularity}
The design distribution has density $p\in C^3\left(\M\right)$ with respect to $\Vol$, and
there are constants $0<p_{\min}\le p_{\max}<\infty$ such that $
  p_{\min}\le p\left(x\right)\le p_{\max}, x\in\M.$
The regression field satisfies $V\in C^3\left(\Gamma\left(T\M\right)\right)$.
\end{assumption}

 \begin{assumption}\label{ASS:Errors}
Choose an everywhere-defined Borel version of the conditional law of $\eta$
given $X=x$.  Its conditional mean is zero, and its covariance field
$\Sigma$ is a self-adjoint $C^3$ section of $\operatorname{End}\left(T\M\right)$.  There
are constants $C_\eta<\infty$ and
$0<\lambda_{\min}\le\lambda_{\max}<\infty$ such that, for every $x\in\M$,
every unit $a\in T_x\M$, and every $t\in\R$,
\[
  \E\left\{\exp\left(t\inner{a}{\eta}_x\right)\mid X=x\right\}
  \le \exp\left(C_\eta^2t^2/2\right),
\]
and
\[
  \lambda_{\min}\norm{u}_x^2
  \le\inner{u}{\Sigma\left(x\right)u}_x
  \le\lambda_{\max}\norm{u}_x^2,
  \qquad u\in T_x\M.
\]
\end{assumption}

 \begin{assumption}\label{ASS:Bandwidth}
The deterministic bandwidth $h=h_n$ satisfies
\begin{equation}\label{EQ:Bandwidth}
  h\downarrow0,
  \qquad
  \frac{\left(\log n\right)^5}{nh^d}\longrightarrow0,
  \qquad
  nh^{d+4}\log\left(1/h\right)\longrightarrow0.
\end{equation}
\end{assumption}

The smoothness and tail conditions support uniform inference.
In the scalar-response setting,
\citet{pelletier2006regression} uses a closed manifold, a normalized
compactly supported kernel, positive design density, and twice continuously
differentiable density and regression functions, with bounded responses for
the pointwise expansions.  Our $C^3$ fields and $C^4$ kernel  provide the
additional derivative control needed for covariance whitening and moving
kernel overlaps.  The sub-Gaussian assumption permits unbounded Gaussian
errors but imposes uniform tail control for the high-dimensional maximum
comparison \citep{chernozhuokov2022improved}; ellipticity makes that
whitening stable.  Local covariance nondegeneracy and regularity are also
central in Gaussian extreme-value theory
 \citep{qiaopolonik2018,qiao2021extremes}. These conditions allow smooth spatially varying anisotropic
covariance.  For example, the normalized nonnegative kernel
$K^\star\left(u\right)=c_d\left(1-\norm u^2\right)^5\mathbf1\left\{\norm u\le1\right\}$
satisfies the kernel condition and can be combined with smooth tangent
fields and conditionally Gaussian errors having smooth uniformly positive
covariance.

The last condition in \eqref{EQ:Bandwidth} undersmooths the second-order bias
so that the confidence tube is centered at $V$.  For $h=n^{-\alpha}$, the two
rate conditions hold when
\begin{equation}\label{EQ:PowerBandwidth}
  \frac{1}{d+4}<\alpha<\frac{1}{d}.
\end{equation}

Similar undersmoothing conditions occur in scalar simultaneous regression
inference.  For example, Assumption~(A5) of \citet{cai2021missingbands}
allows $h=n^{-\alpha}$ with $1/5<\alpha<1/3$, a subrange of
\eqref{EQ:PowerBandwidth} when $d=1$.
For scalar regression on Riemannian manifolds, \citet{henryrodriguez2009}
use $nh^d\to\infty$ and $n^{1/(d+4)}h\to0$ for a centered pointwise
normal limit (Assumption~A3 with $\beta=0$ and Theorem~4.1).
For $h=n^{-\alpha}$, these bandwidth conditions give the same range
\eqref{EQ:PowerBandwidth}.

On a flat Euclidean domain, parallel transport is the identity,
$\Theta_x\left(q\right)=1$, and \eqref{EQ:Estimator} is the usual vector-valued
Nadaraya--Watson estimator.  The formulas below then reduce to the familiar
coordinatewise bias and covariance expressions.

\section{Estimation theory}\label{SEC:Estimation}

The estimator admits an exact decomposition into smoothing bias,
random-design fluctuation, and transported noise.  We use it to derive the
uniform bias and covariance expansions required for simultaneous inference.

 Define the centered transported-signal and noise processes in $T_x\M$ by
\begin{align}
 R_{n,h}\left(x\right)
 &=\frac{1}{\sqrt{nh^d}}\sum_{i=1}^nW_h\left(x,X_i\right)
 \left\{\Tau_{X_i\to x}V\left(X_i\right)-V_h\left(x\right)\right\},
 \notag\\
 U_{n,h}\left(x\right)
 &=\frac{1}{\sqrt{nh^d}}\sum_{i=1}^nW_h\left(x,X_i\right)
 \Tau_{X_i\to x}\eta_i.
 \label{EQ:NoiseNumeratorRenewed}
\end{align}
Then, on every sample,
\begin{equation}\label{EQ:ExactDecomposition}
 \sqrt{nh^d}\left\{\wh V_h\left(x\right)-V_h\left(x\right)\right\}
 =\frac{R_{n,h}\left(x\right)+U_{n,h}\left(x\right)
 +\sqrt{nh^d}\left\{\wh p_h\left(x\right)-\wh p_h^\dagger\left(x\right)\right\}V_h\left(x\right)}{\wh p_h^\dagger\left(x\right)}.
\end{equation}
Here $R_{n,h}$ is the fluctuation of the transported signal under random
design, and $U_{n,h}$ is the leading noise term.  The final numerator term
accounts for denominator truncation.  For all sufficiently large $n$, it
vanishes on
$\mathcal E_{p,n}=\left\{\inf_{x\in\M}\wh p_h\left(x\right)\ge p_{\min}/2\right\}$,
whose probability tends to one.

We first calculate the smoothing bias $V_h-V$ in normal coordinates,
then the covariance of $U_{n,h}$.  Empirical-process bounds over a finite
atlas control the random terms uniformly over $\M$.

For a scalar function $f$, $\nabla f$ is its Riemannian gradient and
$\Delta f=\operatorname{div}(\nabla f)$ is its Laplace--Beltrami operator.
For vector fields, $\nabla_U V$ denotes the covariant derivative of $V$
along $U$; in particular, $\nabla_{\nabla p}V$ differentiates $V$ along
the density gradient.  The connection Laplacian is the trace of the
second covariant derivative:
\begin{equation*}
  \Delta^\nabla V\left(x\right)=\sum_{j=1}^d
  \left\{\nabla_{e_j}\nabla_{e_j}V-
  \nabla_{\nabla_{e_j}e_j}V\right\}\left(x\right),
\end{equation*}
where $\left\{e_1,\ldots,e_d\right\}$ is any local orthonormal frame.

 \begin{lemma}\label{LEM:BiasDensity}
Under Assumptions~\ref{ASS:Kernel}--\ref{ASS:Regularity}, uniformly in
$x\in\M$,
\begin{align*}
  p_h\left(x\right)
  &=p\left(x\right)+\frac{h^2\mu_2\left(K\right)}{2}\Delta p\left(x\right)+O\left(h^3\right),
  \\
  V_h\left(x\right)-V\left(x\right)
  &=h^2\mu_2\left(K\right)
  \left\{\frac{1}{2}\Delta^\nabla V\left(x\right)
  +\frac{1}{p\left(x\right)}\nabla_{\nabla p\left(x\right)}V\left(x\right)\right\}+O\left(h^3\right).
\end{align*}
Under Assumption~\ref{ASS:Bandwidth},
\begin{align*}
 \sup_{x\in\M}|\wh p_h\left(x\right)-p\left(x\right)|
 &=O_{\Pp}\left\{h^2+
 \sqrt{\frac{\log\left(1/h\right)}{nh^d}}\right\},
 \\
 \sup_{x\in\M}\norm{R_{n,h}\left(x\right)}_x
 &=O_{\Pp}\left\{h\sqrt{\log\left(1/h\right)}\right\}.
\end{align*}
\end{lemma}

The leading random fluctuation is determined by the covariance of
$U_{n,h}$ in \eqref{EQ:NoiseNumeratorRenewed}:
\begin{equation}\label{EQ:ChRenewed}
 C_h\left(x\right)=\frac{1}{h^d}\E\left[
 W_h\left(x,X\right)^2\Tau_{X\to x}\Sigma\left(X\right)\Tau_{x\to X}
 \right].
\end{equation}
Its conditional-design analogue, still involving the unknown $\Sigma$, is
\begin{equation*}
 \wh C_h\left(x\right)=\frac{1}{nh^d}\sum_{i=1}^nW_h\left(x,X_i\right)^2
 \Tau_{X_i\to x}\Sigma\left(X_i\right)\Tau_{x\to X_i}.
\end{equation*}
By construction, $
 \wh C_h\left(x\right)=\Cov\left\{U_{n,h}\left(x\right)\mid X_1,\ldots,X_n\right\}. $

The finite-bandwidth population covariance scale for the leading regression
noise and its limit are
\begin{equation*}
  \Omega_h\left(x\right)=\frac{C_h\left(x\right)}{p_h\left(x\right)^2},
  \qquad
  \Omega\left(x\right)=\frac{R\left(K\right)}{p\left(x\right)}\Sigma\left(x\right).
\end{equation*}

 \begin{lemma}\label{LEM:VarianceOperator}
Under Assumptions~\ref{ASS:Kernel}--\ref{ASS:Errors}, uniformly in $x$,
 \[
 \sup_{x\in\M}\left\|\wh C_h\left(x\right)-C_h\left(x\right)\right\|_{\mathrm{op}}
 =O_{\Pp}\left\{\sqrt{\frac{\log\left(1/h\right)}{nh^d}}
                    +\frac{\log\left(1/h\right)}{nh^d}\right\}.
\]
Moreover,
\begin{equation}\label{EQ:VarianceOperatorExpansion}
 C_h\left(x\right)=p\left(x\right)R\left(K\right)\Sigma\left(x\right)+O\left(h^2\right),
\end{equation}
and
\begin{equation}\label{EQ:OmegaExpansion}
 \sup_{x\in\M}\left\|
 \Omega\left(x\right)^{-1/2}\left\{\Omega_h\left(x\right)-\Omega\left(x\right)\right\}\Omega\left(x\right)^{-1/2}
 \right\|_{\mathrm{op}}=O\left(h^2\right).
\end{equation}
\end{lemma}

The volume correction removes the volume-density term from the leading bias,
leaving the connection Laplacian and the interaction between the design
density and the field.  The covariance expansion instead retains one inverse
volume factor because the kernel weight is squared.  Its leading term
$R\left(K\right)\Sigma/p$ determines the local shape of the confidence tube.
 Combining these expansions with \eqref{EQ:ExactDecomposition} gives the uniform
rate.

\begin{theorem}[Uniform rate]\label{THM:UniformRate}
Under Assumptions~\ref{ASS:Kernel}--\ref{ASS:Bandwidth},
\begin{equation*}
 \sup_{x\in\M}\norm{\wh V_h\left(x\right)-V\left(x\right)}_x
 =O_{\Pp}\left\{h^2+
 \sqrt{\frac{\log\left(1/h\right)}{nh^d}}\right\}.
\end{equation*}
\end{theorem}

The stochastic term has the usual effective-sample-size scale $nh^d$, with a
logarithmic factor for uniformity over the manifold.  For inference, this rate
alone is not sufficient: the maximum fluctuates on the smaller Gumbel scale.
Section~\ref{SEC:Inference} therefore retains the leading noise field and
controls the remaining terms separately at that scale.

In a flat chart, away from a boundary, the leading bias is
\begingroup
\[
 h^2\mu_2\left(K\right)\left\{\tfrac12\Delta V+
                   \sum_{j=1}^d\left(\partial_j\log p\right)\partial_jV\right\},
\]
\endgroup
and the covariance is $R\left(K\right)\Sigma/p$: these are the vector-valued
Nadaraya--Watson expressions.  The same formulas hold globally on a flat
torus.  Thus curvature does not change the powers $h^2$ and $\left(nh^d\right)^{-1/2}$;
it enters transport, the volume correction and higher-order terms.
This agrees with the Euclidean-order bias, variance and integrated rates
for scalar manifold regression in \citet{pelletier2006regression}.

\section{Simultaneous inference}\label{SEC:Inference}
Simultaneous inference requires a critical value for the largest
standardized estimation error.  We obtain it in three steps: approximate
the regression maximum by a Gaussian maximum, derive its Gumbel law, and
show that replacing the unknown covariance by an estimate preserves that law.

\subsection{Gaussian reference field and distributional comparison}
\label{SEC:GaussianSection}
To analyze the maximum, we express a vector norm as a scalar supremum:
for every tangent field $G$,
\begin{equation}\label{EQ:SphereBundleReduction}
  \sup_{x\in\M}\norm{G\left(x\right)}_x
  =\sup_{\substack{x\in\M,\ v\in T_x\M\\ \norm{v}_x=1}}
  \inner{v}{G\left(x\right)}_x.
\end{equation}
The unit tangent bundle $S\left(T\M\right)$ was defined in Section~\ref{SEC:Setup}.
Equation~\eqref{EQ:SphereBundleReduction} converts the vector-norm problem to
a scalar supremum without choosing a global frame.

Apply \eqref{EQ:SphereBundleReduction} to the standardized noise
field $G\left(x\right)=C_h\left(x\right)^{-1/2}U_{n,h}\left(x\right)$,
where $U_{n,h}$ and $C_h$ are defined in \eqref{EQ:NoiseNumeratorRenewed}
and \eqref{EQ:ChRenewed}.  Its directional coordinate is
\[
 Z_{n,h}\left(x,v\right)
 =\inner{v}{C_h\left(x\right)^{-1/2}U_{n,h}\left(x\right)}_x,
 \qquad \left(x,v\right)\in S\left(T\M\right).
\]
By \eqref{EQ:NoiseNumeratorRenewed}, this coordinate is a weighted sum of
transported errors.  Its
covariance can be written as an $L^2$ inner product of the corresponding
kernel, transport, and covariance features.  We therefore place all such
features in one Hilbert space and apply a single isonormal Gaussian process.
This construction produces a scalar Gaussian field on $S\left(T\M\right)$ whose
finite-bandwidth covariance agrees exactly with that of the whitened empirical
noise.  All quantities are defined intrinsically on the tangent bundle.

Let $\mathscr H=L^2\left\{\M,T\M,p\,\dd\Vol\right\}$
be the Hilbert space of square-integrable measurable tangent
sections.  An element is a field $U$ assigning $U\left(q\right)\in T_q\M$
to almost every $q$, with
$\int_\M\norm{U\left(q\right)}_q^2p\left(q\right)\dd\Vol\left(q\right)<\infty$;
fields equal almost everywhere under $p\,\dd\Vol$ represent the same element.
For $U,V\in\mathscr H$, the inner product integrates the tangent-space inner products:
\[
 \inner{U}{V}_{\mathscr H}
 =\int_\M\inner{U\left(q\right)}{V\left(q\right)}_q
              p\left(q\right)\dd\Vol\left(q\right).
\]
  For $x\in\M$, define the feature
operator $F_{h,x}:T_x\M\to\mathscr H$ by
\begin{equation}\label{EQ:FhRenewed}
 \left(F_{h,x}v\right)\left(q\right)=h^{-d/2}W_h\left(x,q\right)\Sigma\left(q\right)^{1/2}
 \Tau_{x\to q}C_h\left(x\right)^{-1/2}v.
\end{equation}
The factors in \eqref{EQ:FhRenewed} have distinct roles.  Starting with a
direction $v$ at $x$, the operator $C_h\left(x\right)^{-1/2}$ standardizes the noise
variance, parallel transport moves that direction to a possible observation
location $q$, and $\Sigma\left(q\right)^{1/2}$ accounts for the error covariance there.
The kernel weight specifies how strongly that observation contributes.
Inner products of these features reproduce the covariance between
standardized directional errors at different locations.

Here $F_{h,x}^*:\mathscr H\to T_x\M$ denotes the Hilbert-space adjoint,
defined by
\[
 \inner{F_{h,x}v}{U}_{\mathscr H}=\inner{v}{F_{h,x}^*U}_x,
 \qquad v\in T_x\M,\quad U\in\mathscr H.
\]
Using the definition of $C_h$ gives
$\inner{F_{h,x}v}{F_{h,x}w}_{\mathscr H}=\inner v w_x$.
Equivalently, $F_{h,x}^*F_{h,x}=I_{T_x\M}$, so the feature map is an
exact isometry.  Let $\mathbb W$ be an isonormal Gaussian process over $\mathscr H$: this is
a centered jointly Gaussian family indexed by $U\in\mathscr H$, linear in
$U$, with
$\E\left\{\mathbb W\left(U\right)\mathbb W\left(V\right)\right\}=\inner{U}{V}_{\mathscr H}$.
A single such process supplies the randomness at every location.  Define
\begin{equation*}
  Z_h\left(x,v\right)=\mathbb W\left(F_{h,x}v\right),
  \qquad \left(x,v\right)\in S\left(T\M\right).
\end{equation*}
This field has variance one and exactly the covariance of
the empirical coordinate $Z_{n,h}$ defined above.

The scalar field also defines a Gaussian random section of $T\M$.
For a local orthonormal frame $e_1\left(x\right),\ldots,e_d\left(x\right)$, set
\begin{equation}\label{EQ:GaussianSection}
 \mathcal Z_h\left(x\right)=\sum_{j=1}^d\mathbb W\left\{F_{h,x}e_j\left(x\right)\right\}e_j\left(x\right),
 \qquad
 Z_h\left(x,v\right)=\inner v{\mathcal Z_h\left(x\right)}_x .
\end{equation}
An orthogonal change of local frame changes the Gaussian coordinates and
basis vectors by inverse transformations, leaving the sum unchanged.
The local definitions therefore agree on chart overlaps.  Feature regularity
gives a continuous version, so $\mathcal Z_h$ is a random element of the
space of continuous sections.  It is indexed by $x\in\M$ and takes values in
$T_x\M$; $Z_h$ is its scalar representation indexed by $\left(x,v\right)\in S\left(T\M\right)$.

We next quantify how the estimator inherits the Gaussian calibration.
The relevant normalization is
\begin{equation}\label{EQ:ahRenewed}
  a_h=\sqrt{2d\log\left(1/h\right)}
\end{equation}
because the Gaussian maximum is of order $a_h$, while its fluctuations are
of order $a_h^{-1}$.  Accordingly, an additive perturbation must be
$o_{\Pp}\left(a_h^{-1}\right)$, and a relative covariance perturbation must be
$o_{\Pp}\left(a_h^{-2}\right)$.  The Gaussian comparison itself is distributional.

For the empirical coordinate $Z_{n,h}$, independence and
conditional centering of the observations give the exact covariance identity
\[
 \Cov\left\{Z_{n,h}\left(x,v\right),Z_{n,h}\left(y,w\right)\right\}
 =\inner{F_{h,x}v}{F_{h,y}w}_{\mathscr H}
 =\Cov\left\{Z_h\left(x,v\right),Z_h\left(y,w\right)\right\}.
\]
In particular, the cross-covariance operator of the Gaussian section is
$F_{h,x}^*F_{h,y}:T_y\M\to T_x\M$; at $x=y$ it is the identity.
Consequently $\Omega_h\left(x\right)^{1/2}\mathcal Z_h\left(x\right)$ has the covariance of
$p_h\left(x\right)^{-1}U_{n,h}\left(x\right)$, the leading scaled regression noise.
This covariance match follows from the sampling model and does not require
Gaussian observations.  The remaining issue is whether it also yields an
approximation to the distribution of the maximum.

The target-centered oracle regression maximum is
\begin{equation*}
 T_{n,h}^{V}
 =\sup_{x\in\M}\norm{\Omega\left(x\right)^{-1/2}\sqrt{nh^d}
       \left\{\wh V_h\left(x\right)-V\left(x\right)\right\}}_x.
\end{equation*}

We compare both fields on the same deterministic grid and control the error from
replacing the continuum by that grid.  In a finite bundle atlas, use base
mesh $ha_h^{-3}$ and fibre mesh $a_h^{-3}$.  The grid $\mathcal G_h$ has
$|\mathcal G_h|\lesssim h^{-d}a_h^{3\left(2d-1\right)}$ points.  Uniform scaled first
derivatives of both fields are $O_{\Pp}\left(a_h\right)$, so interpolation changes either
maximum by $O_{\Pp}\left(a_h^{-2}\right)$, smaller than the fluctuation scale $a_h^{-1}$.

At each grid point, the empirical field is a normalized sum of independent
centered scores.  Their variance is one, their sub-exponential norm is
$O\left(h^{-d/2}\right)$, and their fourth moment is $O\left(h^{-d}\right)$.
Theorem~2.1 of \citet{chernozhuokov2022improved} therefore gives
\[
 \sup_t\left|\Pp\left\{\max_{\mathcal G_h}Z_{n,h}\le t\right\}
 -\Pp\left\{\max_{\mathcal G_h}Z_h\le t\right\}\right|
 \le C\left[\frac{\log^5\left\{n|\mathcal G_h|\right\}}{nh^d}\right]^{1/4}=o\left(1\right).
\]
Gaussian anti-concentration \citep{chernozhukov2015comparison} controls the
threshold shifts needed to return to the continuum.  For example, a shift
of $a_h^{-3/2}$ dominates either interpolation error with probability tending
to one and changes the Gaussian grid cdf by $O\left(a_h^{-1/2}\right)$.

Finally, the exact decomposition in Section~\ref{SEC:Estimation} returns us
from the noise maximum to the regression estimator.  With $L_h=\log\left(1/h\right)$,
\[
\begin{split}
 a_h\left|T_{n,h}^{V}-\sup_{S\left(T\M\right)}Z_{n,h}\right|
 &=O_{\Pp}\left[L_h\left\{h^2+\sqrt{\frac{L_h}{nh^d}}
                    +\frac{L_h}{nh^d}\right\}
             +hL_h+\sqrt{nh^{d+4}L_h}\right]=o_{\Pp}\left(1\right).
\end{split}
\]
The final term is the smoothing bias.  The other terms arise from
normalization and transported-signal variation under random design.
Anti-concentration also transfers this small estimator perturbation to a
cdf comparison.  The supplementary material supplies these
arguments, yielding the following proposition.

 \begin{proposition}
\label{PROP:GaussianApproximation}
Under Assumptions~\ref{ASS:Kernel}--\ref{ASS:Bandwidth},
\begin{equation*}
 \sup_{t\in\R}\left|
 \Pp\left\{\sup_{S\left(T\M\right)}Z_{n,h}\le t\right\}-\Pp\left\{\sup_{S\left(T\M\right)}Z_h\le t\right\}\right|\longrightarrow0,
\end{equation*}
and
\begin{equation*}
 a_h\left\lvert T_{n,h}^{V}-\sup_{S\left(T\M\right)}Z_{n,h}\right\rvert=o_{\Pp}\left(1\right).
\end{equation*}
Consequently,
\[
 \sup_{t\in\R}\left|
 \Pp\left\{T_{n,h}^{V}\le t\right\}-\Pp\left\{\sup_{S\left(T\M\right)}Z_h\le t\right\}\right|\longrightarrow0.
\]
\end{proposition}
It remains to find a critical value for the Gaussian maximum.
The next subsection derives its Gumbel limit from the local covariance
structure; Proposition~\ref{PROP:GaussianApproximation} then transfers
this calibration to the regression maximum.

\subsection{Local covariance and Gaussian calibration}
\label{SEC:LocalGaussianGeometry}
The Gaussian maximum depends on how quickly nearby indices decorrelate,
not just on their unit marginal variances.  Moving a location changes the
kernel neighborhood; rotating a tangent direction changes which component
of the vector is measured.  Both changes enter the same Taylor expansion.
For $r_h\left\{\left(x,v\right),\left(y,w\right)\right\}=\Cov\left\{Z_h\left(x,v\right),Z_h\left(y,w\right)\right\}$, exact whitening gives
\begin{equation}\label{EQ:HilbertDistanceIdentity}
 1-r_h\left\{\left(x,v\right),\left(y,w\right)\right\}
 =\frac{1}{2}\norm{F_{h,x}v-F_{h,y}w}_{\mathscr H}^2.
\end{equation}
Thus the quadratic covariance expansion is obtained by differentiating the
feature map and taking its Hilbert-space Gram matrix.

The two kinds of displacement have different scales.  A spatial displacement
of size $h$ changes the translated kernel by order one, whereas the fibre
sphere is not shrinking.  Using $R\left(K\right)$ from
\eqref{EQ:KernelConstants}, define
\begin{equation}\label{EQ:cKRenewed}
 c_K=\frac{\int_{\R^d}\left(\partial_1K^\star\right)^2\dd u}{2R\left(K\right)}>0.
\end{equation}
At a center $s=\left(x,v\right)$, choose orthonormal base and fibre coordinates and
write the bundle chart directly as $\Phi_{h,s}\left(hu,z\right)$, where
$u\in\R^d$ and $z\in\R^{d-1}$.  Here $hu$ is the physical base
coordinate.  The fibre coordinate is adjusted linearly as the base point
moves so that the two feature-derivative blocks are orthogonal at $s$.

Under Assumptions~\ref{ASS:Kernel}--\ref{ASS:Errors}, the joint expansion is
\begingroup
 \begin{equation}\label{EQ:RelativeCovarianceRenewed}
\begin{split}
 &1-r_h\left\{\Phi_{h,s}\left(hu,z\right),\Phi_{h,s}\left(hu',z'\right)\right\}\\
 &\quad=
 \begin{pmatrix}u-u'\\z-z'\end{pmatrix}^{\!\top}
 \begin{pmatrix}c_K I_d+E_{h,s}&0\\0&\tfrac12 I_{d-1}\end{pmatrix}
 \begin{pmatrix}u-u'\\z-z'\end{pmatrix}
 +R_{h,s}\left(u,z,u',z'\right),
\end{split}
\end{equation}
where $\sup_s\norm{E_{h,s}}_{\rm op}\le Ch$ and, on a fixed sufficiently small rescaled coordinate neighborhood,
\[
 |R_{h,s}\left(u,z,u',z'\right)|
 \le C\left(\norm u+\norm{u'}+\norm z+\norm{z'}\right)
             \left\{\norm{u-u'}^2+\norm{z-z'}^2\right\}.
\]
The constants and neighborhood are uniform in $s$ and sufficiently small
$h$.  The spatial derivative block includes a factor $h$ by the chain rule;
the fibre derivative block does not.  The factor $1/2$ multiplying the
Hilbert squared distance accounts for the fibre block $I_{d-1}/2$.
The mixed block is exactly zero at the chart center after the adjustment.
It need not vanish away from the center: its variation is included in the
uniform remainder.
\endgroup

\begingroup
For excursion probabilities, we need this expansion uniformly around every
nearby reference point, not only at the chart center.  For $s=\left(x,v\right)$ and
rescaled coordinates $t=\left(t_u,t_z\right)$, define the $\left(2d-1\right)\times\left(2d-1\right)$ matrix
directly from the correlation function:
\[
 \left[Q_{h,s}\left(t\right)\right]_{ab}
 =\left.\frac12\frac{\partial^2}{\partial t_a\partial t'_b}
 r_h\left\{\Phi_{h,s}\left(ht_u,t_z\right),\Phi_{h,s}\left(ht'_u,t'_z\right)\right\}
 \right|_{t'=t}.
\]
Equivalently, this is one half of the Gram matrix of the derivatives of
$F_{h,x}v$ in these coordinates.  For a small coordinate displacement
$\Delta$, the quadratic form $\Delta^\top Q_{h,s}\left(t\right)\Delta$ gives the
leading loss of correlation.  A larger value means that the Gaussian
field decorrelates more quickly in that direction.  The following statement
gives the uniform control needed to use this calculation throughout the bundle.

\begin{lemma}\label{LEM:LocalCovarianceMain}
Under Assumptions~\ref{ASS:Kernel}--\ref{ASS:Errors}, there are constants
$r_0,c,C>0$, independent of $s$ and sufficiently small $h$, such that the
adjusted charts above are defined on a common rescaled ball
$B_{r_0}\subset\R^{2d-1}$.  For $t,t'\in B_{r_0}$,
\begin{equation}\label{EQ:LocalCovarianceRegularityMain}
 cI_{2d-1}\preceq Q_{h,s}\left(t\right)\preceq CI_{2d-1},
 \qquad
 \norm{Q_{h,s}\left(t\right)-Q_{h,s}\left(t'\right)}_{\rm op}\le C\norm{t-t'}.
\end{equation}
At $t=0$, $Q_{h,s}\left(0\right)$ is the block matrix in
\eqref{EQ:RelativeCovarianceRenewed}.  For any $a,b,t\in B_{r_0}$,
\begin{equation}\label{EQ:LocalCovarianceExpansionMain}
\begin{split}
 &\left|1-r_h\left\{\Phi_{h,s}\left(ha_u,a_z\right),\Phi_{h,s}\left(hb_u,b_z\right)\right\}
            -\left(a-b\right)^\top Q_{h,s}\left(t\right)\left(a-b\right)\right|\\
 &\qquad\le C\max\left\{\norm{a-t},\norm{b-t}\right\}\norm{a-b}^2.
\end{split}
\end{equation}
\end{lemma}

The lower bound in \eqref{EQ:LocalCovarianceRegularityMain} ensures that
every base and fibre direction contributes to the covariance loss.  The
Lipschitz bound and \eqref{EQ:LocalCovarianceExpansionMain} allow the exact
matrix at any reference point to replace nearby matrices with a uniformly
smaller-order error.  At the center, replacing $c_KI_d+E_{h,s}$ by $c_KI_d$
adds at most $Ch\norm{a_u-b_u}^2$.  The supplementary material constructs the adjustment and proves
Lemma~\ref{LEM:LocalCovarianceMain}.
\endgroup

Compact kernel support gives $r_h\left\{\left(x,v\right),\left(y,w\right)\right\}=0$ when $\rho\left(x,y\right)>2h$.
Away from the rescaled diagonal there is also a uniform one-sided gap below
one.  The qualification ``one-sided'' matters because opposite directions
in the same fibre have correlation $-1$.  The exact covariance depends on
$p$ and $\Sigma$, but whitening removes them from the leading quadratic
blocks.  This explains why the leading excursion constant depends on the
kernel, dimension and Riemannian volume.

\label{SEC:GaussianCalibration}
To turn the covariance expansion into an extreme-value law, we use the
local-stationarity ideas developed for manifold-indexed fields by
\citet{qiaopolonik2018,qiao2021extremes}, with the spatial and fibre scales
verified here on the unit tangent bundle.

\begingroup
We first calculate the probability of a high excursion within one small
spatial neighborhood, allowing all tangent directions there.  The local
coefficient is $\sqrt{\det Q_{h,s}\left(t\right)}$: rapid decorrelation in more
directions increases the contribution to the excursion probability.
To combine these coefficients over a region $A\subset S\left(T\M\right)$, partition
$A$, up to chart boundaries, into disjoint pieces represented by coordinate
domains $D_\nu$ in the rescaled charts centered at $s_\nu$, and set
\[
 \mathcal I_h\left(A\right)=h^d\sum_\nu\int_{D_\nu}
             \sqrt{\det Q_{h,s_\nu}\left(t\right)}\dd t.
\]
The determinant transforms with the coordinate Jacobian, so the sum does
not depend on the chosen partition or charts.  The factor $h^d$ compensates
for rescaling the base coordinates.  Thus $\mathcal I_h$ records the
integrated local excursion coefficient using the covariance expansion
already obtained.

Partition $\M$ into $m_h\asymp h^{-d}$ regular cells $J_{k,h}$ of diameter
comparable to $h$, and let $\pi\left(x,v\right)=x$ be the bundle projection.
For a fixed small $\delta>0$, shrink each cell by a
fraction $\delta$ in its simplex coordinates and lift it to
$E_{k,h}^{\delta}=\pi^{-1}\left(J_{k,h}^{\delta}\right)$, including the full fibre
sphere.  The removed boundary strips are the collars.  Let
$\mathscr D_h^{\delta}$ consist of these lifted shrunken cells and the
lifted pieces of a regular triangulation of their collars.
The supplementary material gives a construction with uniform shape and
boundary bounds after rescaling base coordinates by $h^{-1}$.

\begin{proposition}\label{PROP:UniformBundlePickands}
Under Assumptions~\ref{ASS:Kernel}--\ref{ASS:Errors}, fix a sufficiently
small $\delta>0$.  As $u\to\infty$, uniformly over
$A_h\in\mathscr D_h^{\delta}$ and small $h$,
\begin{equation}\label{EQ:OneCellPickandsRenewed}
 \Pp\left\{\sup_{\left(x,v\right)\in A_h}Z_h\left(x,v\right)>u\right\}
 =\left\{1+o\left(1\right)\right\}\pi^{-\left(2d-1\right)/2}u^{2d-1}\Psi\left(u\right)
 h^{-d}\mathcal I_h\left(A_h\right),
\end{equation}
where $\Psi\left(u\right)=u^{-1}\left(2\pi\right)^{-1/2}e^{-u^2/2}$.
\end{proposition}

Proposition~\ref{PROP:UniformBundlePickands} turns the local covariance
geometry into a cell exceedance probability.  Its uniform relative error
is essential because the number of cells diverges as $h\downarrow0$.
To obtain this probability, we partition a rescaled chart into small
blocks and sum their exceedance probabilities.  This sum can count the
same high excursion in several blocks; a matching Bonferroni lower bound
therefore requires control of joint exceedances in pairs of blocks.
We separate the retained block interiors by narrow gaps.  The quadratic
covariance bounds verify the hypotheses of
\citet[Theorem~3.1]{debickihashorvaliu2017}, giving a uniform joint-tail
bound with a factor $e^{-cb^2}$ when the gap is $b/u$ in rescaled
coordinates.  Choosing gaps that grow on the $u^{-1}$ scale but occupy a
vanishing fraction of each block makes the summed pair corrections
negligible.  The correlation gap and a Gaussian concentration bound
handle pairs at fixed positive separation.  The supplementary material
gives these bounds and the boundary-strip calculation.
The center blocks in \eqref{EQ:RelativeCovarianceRenewed} also give
\[
 \mathcal I_h\left\{S\left(T\M\right)\right\}
 \longrightarrow c_K^{d/2}2^{-\left(d-1\right)/2}\Vol\left(\M\right)\Vol\left(S^{d-1}\right).
\]
Consequently, summing \eqref{EQ:OneCellPickandsRenewed} over the manifold
produces an intensity proportional to $h^{-d}u^{2d-2}e^{-u^2/2}$.
\endgroup

With $a_h$ as in \eqref{EQ:ahRenewed}, put
\begin{equation}\label{EQ:CvecRenewed}
 C_{\mathrm{vec}}
 =\frac{2^{d/2}d^{d-1}}{\pi^{d/2}\Gamma\left(d/2\right)}
 c_K^{d/2}\Vol\left(\M\right).
\end{equation}
The constant $C_{\mathrm{vec}}$ combines the base-volume and fibre-area
integrals above.
The classical expanded Gumbel centering is
\begin{equation}\label{EQ:BetaVectorRenewed}
 \beta_h^{\mathrm{vec}}
 =a_h+\frac{\left(d-1\right)\log\log\left(1/h\right)+\log C_{\mathrm{vec}}}{a_h}.
\end{equation}

\begingroup
To see why this intensity gives a Gumbel law, fix $z\in\R$ and set
$u_{h,z}=\beta_h^{\mathrm{vec}}+z/a_h$.  Count the shrunken cells that
contain an exceedance:
\[
 W_{h,\delta}=\sum_{k=1}^{m_h}
 \1\left\{\sup_{E_{k,h}^{\delta}}Z_h>u_{h,z}\right\},
 \qquad \lambda_{h,\delta}=\E W_{h,\delta}.
\]
Summing the uniform cell probabilities gives the mean of this count.
Cells more than $2h$ apart have independent Gaussian fields, so each cell
depends on only a bounded number of neighbors.  The correlation gap makes
joint exceedances in neighboring shrunken cells negligible.  The
Chen--Stein bound of \citet{arratiagoldsteingordon1989} then yields, for
fixed $\delta$ and $z$ as $h\downarrow0$,
\begin{equation}\label{EQ:PoissonAggregationMain}
\begin{split}
 &\lambda_{h,\delta}=e^{-z}+O\left(\delta\right)+o\left(1\right),\\
 &d_{\rm TV}\left\{\mathcal L\left(W_{h,\delta}\right),
                   \operatorname{Po}\left(\lambda_{h,\delta}\right)\right\}\longrightarrow0.
\end{split}
\end{equation}
Here $d_{\rm TV}$ is total variation distance and $\operatorname{Po}\left(\lambda\right)$
denotes the Poisson law with mean $\lambda$.  The probability of an
exceedance in the omitted collars is $O\left(\delta\right)+o\left(1\right)$.  Thus
\eqref{EQ:PoissonAggregationMain} converts the event of no exceedance into
\[
 \Pp\left\{\sup_{S\left(T\M\right)}Z_h\le u_{h,z}\right\}
 =\exp\left\{-e^{-z}\right\}+O\left(\delta\right)+o\left(1\right).
\]
Letting $h\downarrow0$ first and then $\delta\downarrow0$ proves the
following theorem.  The supplementary material supplies the joint-tail bounds, Poisson
approximation and collar calculation.
\endgroup

\begin{theorem}[Gaussian sphere-bundle Gumbel law]
\label{THM:GaussianBundle}
Under Assumptions~\ref{ASS:Kernel}--\ref{ASS:Errors}, as $h\downarrow0$ with
$h<\iota_\M/2$, for every $z\in\R$,
\begin{equation}\label{EQ:GaussianBundleGumbelRenewed}
 \Pp\left[a_h\left\{
 \sup_{\left(x,v\right)\in S\left(T\M\right)}Z_h\left(x,v\right)-\beta_h^{\mathrm{vec}}
 \right\}\le z\right]
 \longrightarrow\exp\left\{-e^{-z}\right\}.
\end{equation}
The full fibre sphere contains both scalar signs, and its area therefore
supplies the intensity for the tangent norm.
\end{theorem}

{
Combining Proposition~\ref{PROP:GaussianApproximation} with
Theorem~\ref{THM:GaussianBundle} gives, for every $z\in\R$,
\[
 \Pp\left[a_h\left\{T_{n,h}^{V}-\beta_h^{\mathrm{vec}}\right\}\le z\right]
 \longrightarrow \exp\{-e^{-z}\}.
\]
On a flat torus, this is the joint norm limit for periodic
$\R^d$-valued kernel regression. The corresponding Euclidean result holds
on a compact inference region inside the design support.

\begin{proposition}
\label{PROP:EuclideanSpecialization}
Let $d\ge1$, $D=\prod_{j=1}^d[\ell_j,r_j]$ with $\ell_j<r_j$, and consider
independent copies of $(X,Y)$ satisfying
$Y=V(X)+\eta$, where $X,Y\in\R^d$.
Suppose Assumptions~\ref{ASS:Kernel} and~\ref{ASS:Bandwidth} hold, and the
Euclidean versions of Assumptions~\ref{ASS:Regularity}--\ref{ASS:Errors}
hold uniformly on a fixed open neighborhood of $D$.
Use the Euclidean estimator~\eqref{EQ:Estimator}, retaining observations
outside $D$. With $|D|$ denoting Lebesgue volume, set
\begin{align*}
 T_{n,h,D}^{V}
 &=\sup_{x\in D}\left\|\Omega(x)^{-1/2}\sqrt{nh^d}
       \{\wh V_h(x)-V(x)\}\right\|,\qquad
 C_D=\frac{2^{d/2}d^{d-1}}{\pi^{d/2}\Gamma(d/2)}c_K^{d/2}|D|,\\
 \beta_{h,D}
 &=a_h+\frac{(d-1)\log\log(1/h)+\log C_D}{a_h},
 \qquad a_h=\sqrt{2d\log(1/h)}.
\end{align*}
Then, for every $z\in\R$,
\begin{equation}\label{EQ:EuclideanGumbel}
 \Pp\left[a_h\{T_{n,h,D}^{V}-\beta_{h,D}\}\le z\right]
 \longrightarrow\exp\{-e^{-z}\}.
\end{equation}
The limit also holds with $\Omega$ replaced by a continuous positive-definite
estimate whose uniform relative error is $o_{\Pp}(a_h^{-2})$, as in
\eqref{EQ:RelativeStudentizationRenewed} with the supremum taken over $D$.
\end{proposition}

For $d=1$, write $D=[\ell,r]$, $L=r-\ell$ and
$\Sigma(x)=\sigma^2(x)$. Then
\begin{align*}
 T_{n,h,D}^{V}
 &=\sup_{x\in D}\frac{\sqrt{nh\,p(x)}}{\sqrt{R(K)}\,\sigma(x)}
       |\wh V_h(x)-V(x)|,\qquad
 C_D=\frac{L\sqrt{2c_K}}{\pi}.
\end{align*}
There is no $\log\log(1/h)$ term. Centering instead at
$\beta_{h,D}^{+}=\beta_{h,D}-(\log2)/a_h$ gives the limiting cdf
$\exp\{-2e^{-z}\}$. This agrees with the smooth-kernel calibration of
\citet[Theorem~2.4]{liuwu2010simultaneous} under independence and negligible
bias: their $\lambda_K$ and $K_2$ equal $R(K)$ and $c_K$, and their rescaled
bandwidth is $h/L$. Here $K^\star(\pm1)=0$.
Inverting the studentized limit gives the usual simultaneous interval band.

For $d\ge2$, $C_D$ agrees with the smooth $\chi_d$-field normalization in
\citet[Corollary~3.1]{qiao2021extremes}. When $p$ and $\Sigma$ are constant
near $D$, the Gaussian section has independent stationary coordinates;
local whitening gives the same leading constant when they vary.
\citet{proksch2016bands} constructs nonparametric bands for a scalar
response with multidimensional fixed design, rather than a joint
vector-norm tube. \citet{liuetal2016tubes} construct exact simultaneous
confidence tubes for Gaussian multivariate linear regression with a common
error covariance. Their ellipsoidal cross-sections have the same form as
our Euclidean tube, but their critical value is obtained from a
largest-root distribution. That finite-sample calibration is not a
special case of the shrinking-bandwidth Gumbel law.
The proof of Proposition~\ref{PROP:EuclideanSpecialization} is given in the supplementary material.  Returning to $\M$, we now estimate
$\Omega$ at the required relative accuracy.
}

\subsection{Covariance estimation}
\label{SEC:CovarianceEstimation}

The oracle statistic contains the unknown covariance through
$\Omega\left(x\right)=R\left(K\right)\Sigma\left(x\right)/p\left(x\right)$.  Directly
averaging residual outer products with the undersmoothed inference bandwidth
$h$ can be unstable even when the numerator already has enough local
observations for Gaussian approximation.  We therefore use a mean-pilot
bandwidth $b$ and a covariance bandwidth $g$, neither of which is tied to $h$.

The computation has three stages.  We first predict the mean using observations
outside each held-out fold and use the prediction errors as residuals.
Next, we estimate the covariance field by smoothing their transported outer
products at bandwidth $g$ and stabilize small eigenvalues.  Finally, we
aggregate these covariance estimates with the squared weights of the original
regression estimator at bandwidth $h$.

For the first stage, split the sample deterministically into a fixed number $K_0\ge2$ of folds with
sizes $n_k$ such that, for some $c_F>0$ and all large $n$,
\[
 \min_k\frac{n_k}{n}\ge c_F,
 \qquad
 \min_k\frac{n-n_k}{n}\ge c_F.
\]
For fold $k$, fit the estimator in \eqref{EQ:Estimator} at bandwidth $b$
using only observations outside that fold, with the training-sample size in
its normalization.  Denote this fit by $\wh V_b^{\left(-k\right)}$.
If observation $i$ belongs to fold $k\left(i\right)$, define
\begin{equation}\label{EQ:CrossFittedResidualRenewed}
 \wh\eta_{i,b}^{\mathrm{cf}}
 =Y_i-\wh V_b^{\left(-k\left(i\right)\right)}\left(X_i\right)\in T_{X_i}\M.
\end{equation}
Since observation $i$ is not used in its own prediction, the fitted mean is
independent of its held-out error conditional on the training data and $X_i$.
For the second stage, transport residuals from nearby source points into the
same target fibre and average their outer products.  Smoothing at bandwidth
$g$ gives
\begin{equation}\label{EQ:CrossFitRawCovarianceRenewed}
 \wh\Sigma_{b,g}\left(x\right)
 =\frac{1}{ng^d\wh p_g^\dagger\left(x\right)}
 \sum_{i=1}^nW_g\left(x,X_i\right)
 \left\{\Tau_{X_i\to x}\wh\eta_{i,b}^{\mathrm{cf}}\right\}^{\otimes2}.
\end{equation}
Here $\wh p_g^\dagger$ is the floored density estimate from
Section~\ref{SEC:Setup}, evaluated at bandwidth $g$.  All operators in
the average act on $T_x\M$, so the estimate is defined separately at each location.

If few residuals receive appreciable weight, this local matrix can be
singular.  We therefore floor its eigenvalues on a scale that vanishes with
sample size.  Choose the strictly positive intrinsic residual scale
\begin{equation*}
 \wh s_{n,b}^{\mathrm{cf}}
 =\left\{\frac{1}{dn}\sum_{i=1}^n
 \norm{\wh\eta_{i,b}^{\mathrm{cf}}}_{X_i}^2\right\}\vee\underline s_n,
 \qquad \underline s_n=n^{-1},
\end{equation*}
and set $\gamma_n=\left\{\log\left(en\right)\right\}^{-2}$.  For a positive semidefinite operator
$A=\sum_j\lambda_j e_j\otimes e_j$ and a scalar $s>0$, let
\begin{equation}\label{EQ:IntrinsicCovarianceFloor}
 \mathcal F_{\gamma,s}\left(A\right)
 =\sum_{j=1}^d
 \left[\lambda_j\vee
 \gamma\left\{\frac{\operatorname{tr}\left(A\right)}{d}+s\right\}\right]
 e_j\otimes e_j .
\end{equation}
Set $\wh\Sigma_{b,g}^{\mathrm{fl}}
=\mathcal F_{\gamma_n,\wh s_{n,b}^{\mathrm{cf}}}\left(\wh\Sigma_{b,g}\right)$.
The functional calculus in \eqref{EQ:IntrinsicCovarianceFloor} is
frame-invariant.  On every finite sample its smallest eigenvalue is at least
$\gamma_n\underline s_n>0$ and its condition number is at most
$d/\gamma_n$, while Proposition~\ref{PROP:CrossFitStudentizationRenewed}
shows that it is uniformly inactive with probability tending to one.

For the final stage, recall that the covariance of a weighted sum uses
squared weights.  We evaluate the stabilized covariance field at each source
point, transport it to the target, and combine it with the original regression
weights.  Define these weights and the resulting covariance estimate by
\begin{align}
 a_{i,h}^\dagger\left(x\right)
 &=\frac{W_h\left(x,X_i\right)}{nh^d\wh p_h^\dagger\left(x\right)},\notag\\
 \wh\Omega_{n,h\mid g}\left(x\right)
 &=nh^d\sum_{i=1}^n\left\{a_{i,h}^\dagger\left(x\right)\right\}^2
 \Tau_{X_i\to x}\wh\Sigma_{b,g}^{\mathrm{fl}}\left(X_i\right)\Tau_{x\to X_i}.
 \label{EQ:OmegaHat}
\end{align}
On the event that the denominator floor is uniformly inactive,
$a_{i,h}^\dagger=W_h/\sum_jW_h$, and the oracle version
\begin{equation*}
 \Omega_{n,h}^{X}\left(x\right)
 =nh^d\sum_{i=1}^n\left\{a_{i,h}^\dagger\left(x\right)\right\}^2
 \Tau_{X_i\to x}\Sigma\left(X_i\right)\Tau_{x\to X_i}
\end{equation*}
is exactly the conditional covariance of the scaled noise term in
$\wh V_h\left(x\right)$.  Thus \eqref{EQ:OmegaHat} retains both the realized weights
and the spatially varying error covariance.

To state the accuracy needed for this replacement, let $L_u=\log\left(1/u\right)$ and
write the uniform error rate as
$
 r_{n,b}=b^2+\sqrt{n^{-1}b^{-d}\log n}+n^{-1}b^{-d}\log n.$
The bandwidth conditions are
\begin{equation}\label{EQ:TwoBandwidthConditionsMain}
 \begin{gathered}
 b,g\downarrow0,\qquad \frac{nb^d}{\log n}\to\infty,\qquad
 \frac{ng^d}{L_g}\to\infty,\\
 L_h\left\{h^2+\sqrt{\frac{L_h}{nh^d}}+\frac{L_h}{nh^d}
 +g^2+\left(1+r_{n,b}\right)\left(\sqrt{\frac{L_g}{ng^d}}+\frac{L_g}{ng^d}\right)
 +r_{n,b}^2\right\}\to0.
 \end{gathered}
\end{equation}
These conditions allow a wider neighborhood for covariance estimation
than for regression, increasing the information available to estimate
$\Sigma$.  A concrete compatible choice, with $\alpha$ in
\eqref{EQ:PowerBandwidth}, is
\[
 b\asymp g\asymp\left(\log n/n\right)^{1/\left(d+4\right)},
 \qquad h=n^{-\alpha},\qquad
 \frac{1}{d+4}<\alpha<\frac{1}{d}.
\]
Thus the usual smooth-function bandwidths for the nuisance fits are
compatible with the undersmoothed regression bandwidth.  The general
conditions in \eqref{EQ:TwoBandwidthConditionsMain} impose no fixed ratio
between $g$ and $h$.

 \begin{proposition}
\label{PROP:CrossFitStudentizationRenewed}
Under Assumptions~\ref{ASS:Kernel}--\ref{ASS:Bandwidth},
\eqref{EQ:TwoBandwidthConditionsMain}, and the floor sequences specified
above,
\begin{equation*}
 \max_{1\le k\le K_0}\sup_{x\in\M}
 \norm{\wh V_b^{\left(-k\right)}\left(x\right)-V\left(x\right)}_x
 =O_{\Pp}\left(r_{n,b}\right).
\end{equation*}
Moreover, with
\[
 e_{\Sigma,n}
 =\sup_{x\in\M}\norm{\Sigma\left(x\right)^{-1/2}
 \left\{\wh\Sigma_{b,g}^{\mathrm{fl}}\left(x\right)-\Sigma\left(x\right)\right\}
 \Sigma\left(x\right)^{-1/2}}_{\mathrm{op}},
\]
\begin{equation*}
 e_{\Sigma,n}
 =O_{\Pp}\left\{g^2+\left(1+r_{n,b}\right)\left(\sqrt{\frac{L_g}{ng^d}}
 +\frac{L_g}{ng^d}\right)+r_{n,b}^2\right\}.
\end{equation*}
On the event $e_{\Sigma,n}<1$,
\begin{equation}\label{EQ:TwoBandwidthLoewnerMain}
 \left(1-e_{\Sigma,n}\right)\Omega_{n,h}^{X}\left(x\right)
 \preceq\wh\Omega_{n,h\mid g}\left(x\right)
 \preceq\left(1+e_{\Sigma,n}\right)\Omega_{n,h}^{X}\left(x\right)
\end{equation}
simultaneously for every $x$.  Consequently,
\begin{equation*}
 \begin{aligned}
 &\sup_{x\in\M}\norm{\Omega\left(x\right)^{-1/2}
 \left\{\wh\Omega_{n,h\mid g}\left(x\right)-\Omega\left(x\right)\right\}
 \Omega\left(x\right)^{-1/2}}_{\mathrm{op}}\\
 &\quad=O_{\Pp}\left\{h^2+\sqrt{\frac{L_h}{nh^d}}+\frac{L_h}{nh^d}
 +g^2+\left(1+r_{n,b}\right)\left(\sqrt{\frac{L_g}{ng^d}}+\frac{L_g}{ng^d}\right)
 +r_{n,b}^2\right\}.
 \end{aligned}
\end{equation*}
In particular,
\begin{equation}\label{EQ:RelativeStudentizationRenewed}
 L_h\sup_{x\in\M}\norm{\Omega\left(x\right)^{-1/2}
 \left\{\wh\Omega_{n,h\mid g}\left(x\right)-\Omega\left(x\right)\right\}
 \Omega\left(x\right)^{-1/2}}_{\mathrm{op}}
 =o_{\Pp}\left(1\right).
\end{equation}
The spectral floor in \eqref{EQ:IntrinsicCovarianceFloor} is uniformly
inactive with probability tending to one.
\end{proposition}

The Loewner transfer in \eqref{EQ:TwoBandwidthLoewnerMain} is pathwise.  Hence
evaluating $\wh\Sigma_{b,g}^{\mathrm{fl}}$ at the random $X_i$, and reusing
those observations in the $h$-sandwich, requires no additional covariance
sample split.  The factor $L_h\asymp a_h^2$ in
\eqref{EQ:RelativeStudentizationRenewed} is the exact scale required to replace
the oracle inverse square root in a maximum of order $a_h$.

The proof of Proposition~\ref{PROP:CrossFitStudentizationRenewed} is in
the supplementary material.  Cross-fitting makes the
residual--pilot cross term conditionally centered; its remaining contribution
is smaller than the covariance-smoothing error.  The squared pilot error
accounts for the $r_{n,b}^2$ term.  Thus broad covariance neighborhoods can
stabilize studentization while $h$ remains small enough for inference on $V$.

\label{SEC:FeasibleTube}

We now replace the oracle covariance by its estimate and invert the
resulting Gumbel law to obtain a confidence tube.

Let $\underline\omega_n>0$ be deterministic with
$\underline\omega_n\downarrow0$.  For
$A=\sum_j\lambda_je_j\otimes e_j$, write
\begin{equation*}
 \left[A\right]_{\tau}^{-1/2}
 =\sum_j\left(\lambda_j\vee\tau\right)^{-1/2}e_j\otimes e_j.
\end{equation*}
The final clip makes the statistic measurable even when the $h$-denominator
gate fails and is asymptotically inactive.  The feasible statistic is
\begin{equation*}
 \wh T_n=\sup_{x\in\M}
 \norm{\left[\wh\Omega_{n,h\mid g}\left(x\right)\right]_{\underline\omega_n}^{-1/2}
 \sqrt{nh^d}\left\{\wh V_h\left(x\right)-V\left(x\right)\right\}}_x.
\end{equation*}

The preceding relative covariance rate implies $
 a_h|\wh T_n-T_{n,h}^{V}|=o_{\Pp}\left(1\right).
$
Indeed, the Loewner bounds make the norm perturbation at most a constant
times the relative covariance error multiplied by $T_{n,h}^{V}=O_{\Pp}\left(a_h\right)$.
The factor $a_h^2$ in the required rate is therefore essential.

For the feasible law, retain the polynomial factor in the leading excursion
intensity and define the unexpanded centering
\begin{equation}\label{EQ:UnexpandedCenteringMain}
 \lambda_h^{\left(0\right)}\left(u\right)
 =h^{-d}\frac{C_{\mathrm{vec}}}{\left(2d\right)^{d-1}}
   u^{2d-2}e^{-u^2/2},
 \qquad
 \widetilde\beta_h^{\mathrm{vec}}>\sqrt{2d-2},
 \qquad
 \lambda_h^{\left(0\right)}\left(\widetilde\beta_h^{\mathrm{vec}}\right)=1.
\end{equation}
The high-branch solution is unique and exists for all sufficiently small
$h$.

\begin{theorem}[Feasible full-norm Gumbel law]
\label{THM:FullNormGumbel}
Under Assumptions~\ref{ASS:Kernel}--\ref{ASS:Bandwidth} and
\eqref{EQ:TwoBandwidthConditionsMain},
\begin{equation*}
 \sup_{z\in\R}\left|
 \Pp\left[a_h\left\{\wh T_n-\widetilde\beta_h^{\mathrm{vec}}\right\}\le z\right]
 -\exp\left\{-e^{-z}\right\}\right|\longrightarrow0.
\end{equation*}
\end{theorem}

The centering in \eqref{EQ:UnexpandedCenteringMain} changes the
finite-$h$ location but not the limiting law.  In fact, $
 a_h\left\{\widetilde\beta_h^{\mathrm{vec}}
       -\beta_h^{\mathrm{vec}}\right\}=o\left(1\right).
$
The equivalence lemma and its proof are given in the supplementary material.

To turn this limit into confidence regions, take its $\left(1-\alpha\right)$ quantile:
\begin{equation}\label{EQ:GumbelQuantile}
 q_{1-\alpha}=-\log\left\{-\log\left(1-\alpha\right)\right\},
 \qquad
 c_{n,1-\alpha}=\widetilde\beta_h^{\mathrm{vec}}
 +\frac{q_{1-\alpha}}{a_h}.
\end{equation}
The feasible $\left(1-\alpha\right)$ simultaneous confidence tube is
\begin{equation}\label{EQ:ConfidenceTube}
 \mathcal C_{n,1-\alpha}\left(x\right)
 =\left\{u\in T_x\M:
 \sqrt{nh^d}\norm{\left[\wh\Omega_{n,h\mid g}\left(x\right)\right]_{\underline\omega_n}^{-1/2}
 \left\{u-\wh V_h\left(x\right)\right\}}_x\le c_{n,1-\alpha}\right\}.
\end{equation}
Theorem~\ref{THM:FullNormGumbel} implies
\begin{equation}\label{EQ:ConfidenceCoverage}
\Pp\left\{V\left(x\right)\in\mathcal C_{n,1-\alpha}\left(x\right)
       \text{ for every }x\in\M\right\}\longrightarrow1-\alpha.
\end{equation}

At each location, \eqref{EQ:ConfidenceTube} gives an ellipsoid whose
orientation and relative axes come from the estimated covariance;
$c_{n,1-\alpha}$ supplies the common multiplicity correction across the domain.
The proof of Theorem~\ref{THM:FullNormGumbel} is in the supplementary material.

 \begin{algorithm}[h]
\caption{Simultaneous tangent-field inference at supplied bandwidths}
\label{ALG:Tube}
\begin{algorithmic}[1]
\normalsize
\Require Tangent observations, known geometry, kernel, level $1-\alpha$,
 fixed folds, $h,b,g$, and vanishing numerical floors.
\State Compute the volume-corrected $h$-weights, density and ratio estimate
\eqref{EQ:Estimator}.  Record  all unsupported targets.
\State Fit each out-of-fold source pilot at $b$ and form residuals
\eqref{EQ:CrossFittedResidualRenewed}.  Smooth  transported residual outer
products at  $g$ using \eqref{EQ:CrossFitRawCovarianceRenewed}.
 \State Compute the residual scale  with anchor $n^{-1}$  and apply
\eqref{EQ:IntrinsicCovarianceFloor} with $\gamma_n=\left\{\log\left(en\right)\right\}^{-2}$.
 Insert source covariances in the realized squared-weight sandwich
\eqref{EQ:OmegaHat}, then apply the final inverse-square-root clip.
\State Compute $a_h$ and $C_{\mathrm{vec}}$.   Bracket the centering root
$\lambda_h^{\left(0\right)}\left(u\right)=1$ on $u>\sqrt{2d-2}$.
\If{the high root does not exist or a required computation fails}
\State Report the calibration unavailable; do not substitute the low root.
\Else
\State Solve on the decreasing branch and set
$c_{n,1-\alpha}=u+q_{1-\alpha}/a_h$ as in \eqref{EQ:GumbelQuantile}.
Return \eqref{EQ:ConfidenceTube}.
 \EndIf
\end{algorithmic}
\end{algorithm}

The deterministic-bandwidth guarantee requires \eqref{EQ:Bandwidth} and
\eqref{EQ:TwoBandwidthConditionsMain}, with
$h<\min\left(1,\iota_\M/2\right)$ and $b,g<\iota_\M$.
For a smoothed target, the target-local covariance
in Section~\ref{SEC:NCEPWind} replaces the source-covariance steps.
The probability-specific equation
$\lambda_h^{\left(0\right)}\left(c\right)=-\log\left(1-\alpha\right)$ on its high branch gives an
asymptotically equivalent, but not identical finite-sample, cutoff
(see the supplementary material).  The supplementary material gives the safeguards and their coverage conditions.  Numerical
approximation of the continuum maximum preserves the limit when its error
is $o_{\Pp}\left(a_h^{-1}\right)$.

\subsection{A geometry-aware bandwidth family}\label{SEC:UnifiedBandwidth}
\begingroup
The three smoothing steps need different amounts of local information, but
their bandwidths can be organized around one reference radius.  We describe
a common choice for uniform designs on $S^2$, $\mathbb T^2$, $S^3$,
$\mathbb T^3$ and $SO\left(3\right)$, allowing spatially varying covariance.

Let $\omega_d$ be the Euclidean unit-ball volume and set
$\ell_\M=\left\{\Vol\left(\M\right)/\omega_d\right\}^{1/d}$.  Write $\kappa_\M$ for a bound on
absolute sectional curvature.  To keep the kernels inside normal
neighborhoods and limit volume distortion, use
\[
 R_\M=\min\left\{0.9\iota_\M,\pi/\left(2\sqrt{\kappa_\M}\right)\right\},
 \qquad H_\M=\min\left\{0.45\iota_\M,R_\M\right\},
\]
where the curvature bound contributes no cap when $\kappa_\M=0$.
On these manifolds the normal-coordinate volume density is a radial
function $\Theta_\M\left(t\right)$.  The effective fraction of the sample used by a
volume-corrected local mean at radius $r$ is
\begin{equation}\label{EQ:GeometricEffectiveMass}
 J_{2,\M}\left(r\right)=\int_{\|u\|\le1}
       \frac{\left\{K^\star\left(u\right)\right\}^2}{\Theta_\M\left(r\|u\|\right)}\dd u,
 \qquad e_\M\left(r\right)=\frac{r^d}{\Vol\left(\M\right)J_{2,\M}\left(r\right)}.
\end{equation}
Thus $ne_\M\left(r\right)$ in \eqref{EQ:GeometricEffectiveMass} accounts for the unequal kernel weights when
measuring the information available within radius $r$.

Choose the smallest positive solution of
\begin{equation}\label{EQ:UnifiedReferenceRadius}
 ne_\M\left(r_n\right)\left(r_n/\ell_\M\right)^4=1,\qquad 0<r_n\le R_\M,
\end{equation}
and use $r_n=R_\M$ if there is no solution in this range.  The fourth power
represents squared second-order smoothing bias, whereas
$\left\{ne_\M\left(r\right)\right\}^{-1}$ represents variance.  Equation~\eqref{EQ:UnifiedReferenceRadius}
balances these dimensionless reference scales using geometric quantities.
The multiplier below allows their overall scale to be adjusted.
For a positive multiplier $\theta$, set
\begin{equation}\label{EQ:UnifiedBandwidthFamily}
 \begin{aligned}
 h_n\left(\theta\right)&=\min\left\{H_\M,\theta r_n\left\{\log\left(en\right)\right\}^{-2/\left(d+4\right)}\right\},\quad
 b_n\left(\theta\right)&=\min\left\{R_\M,\theta r_n\right\},\\
 g_n\left(\theta\right)&=\min\left\{R_\M,\theta r_n\left\{\log\left(en\right)\right\}^{1/\left(d+4\right)}\right\}.
 \end{aligned}
\end{equation}
The mean is undersmoothed for inference on $V$, the residual pilot uses the
reference scale, and covariance smoothing uses a wider neighborhood.
Dimension determines the powers, volume fixes the length scale, and
curvature enters both the effective count and the geometric cap.

The default is $\theta=1$.  A finite candidate set
$\left\{2^{j/4}:j=-4,\ldots,4\right\}$ allows a factor-four range of reference scales;
out-of-fold prediction loss at $b_n\left(\theta\right)$ can select among supported
candidates.  Here prediction loss is
the mean squared tangent-norm error on held-out observations; empty
neighborhoods must be excluded and local effective counts checked separately.
Since $J_{2,\M}\left(r\right)=R\left(K\right)+O\left(r^2\right)$,
the reference radius has order $n^{-1/\left(d+4\right)}$ and the geometric
caps eventually become inactive.  For deterministic $\theta$ bounded above
and away from zero, the logarithmic factors in
\eqref{EQ:UnifiedBandwidthFamily} then give
$nh_n^{d+4}\log\left(1/h_n\right)=O\left(1/\log n\right)$ and
$nh_n^d/\left(\log n\right)^5\to\infty$; the pilot and covariance
rates also satisfy \eqref{EQ:TwoBandwidthConditionsMain}.
The deterministic choices therefore satisfy the asymptotic inference
conditions.  Coverage after selecting $\theta$ by prediction loss requires
a separate analysis.
\endgroup

\section{Simulation studies}\label{SEC:SimulationDesign}

The theory provides an asymptotic simultaneous guarantee.  We now examine
how its analytic calibration behaves at the available sample sizes, using
five geometries: $S^2$ and $\mathbb T^2$ in dimension two, and $S^3$,
$\mathbb T^3$ and $SO\left(3\right)$ in dimension three.  Spheres and tori contrast
curved and flat domains, while the rotation group represents orientation
data.  All five experiments estimate the covariance from the observations.

\subsection{Common design}

The designs are uniform, the regression fields are smooth, and the errors
are conditionally Gaussian.  All experiments use the kernel
\begin{equation*}
 K^\star\left(u\right)=c_d\left(1-\norm u^2\right)^5\mathbf1\left\{\norm u\le1\right\},
 \qquad c_d=\frac{\Gamma\left(d/2+6\right)}{120\pi^{d/2}}.
\end{equation*}
For each $n\in\left\{200,400,600,800,1000,2000,4000\right\}$, we use $500$ replicates,
five cross-fitting folds and a nominal coverage of $95\%$.
The bandwidths $h,b,g$ were fixed after independent pilot and validation
runs and before the $500$ production replicates.  Residuals and covariance estimates
are computed by the procedure in Section~\ref{SEC:CovarianceEstimation}.
Details of the cutoff calculation and numerical maximization
are given in the supplementary material.

The results below use these fixed schedules; the family in
Section~\ref{SEC:UnifiedBandwidth} is a separate bandwidth construction.

\begingroup
\setlength{\intextsep}{10pt plus 2pt minus 2pt}
\setlength{\textfloatsep}{10pt plus 2pt minus 2pt}
\setlength{\floatsep}{10pt plus 2pt minus 2pt}
\begin{table}[!htbp]
\centering
\caption{Analytic-cutoff frequencies at nominal level $0.95$ with fixed bandwidths,
from $500$ replicates per cell.  Columns identify the manifold and intrinsic
dimension. }
\label{TAB:SimulationResults}
 \begin{tabular}{cccccc}
\toprule
$n$ & $S^2$ & $\mathbb T^2$ & $S^3$ & $\mathbb T^3$ & $SO\left(3\right)$\\
$d$ & $2$ & $2$ & $3$ & $3$ & $3$\\
\midrule
200  & 0.696 & 0.820 & 0.904 & 0.536 & 0.756\\
400  & 0.970 & 0.926 & 0.936 & 0.850 & 0.902\\
600  & 0.982 & 0.942 & 0.964 & 0.910 & 0.936\\
800  & 0.970 & 0.958 & 0.962 & 0.942 & 0.948\\
1000 & 0.956 & 0.970 & 0.956 & 0.944 & 0.960\\
2000 & 0.960 & 0.960 & 0.974 & 0.964 & 0.966\\
4000 & 0.956 & 0.950 & 0.962 & 0.956 & 0.942\\
\bottomrule
\end{tabular}
\end{table}
The analytic calibration stabilizes as the sample size increases
(Table~\ref{TAB:SimulationResults}).  For $n\ge1000$, frequencies range
from $0.942$ to $0.974$, close to the nominal $0.95$ on the reported
Monte Carlo scale.  At $n=200$, the much lower frequencies in several
geometries show the difficulty of simultaneous calibration with sparse
local information.  All comparisons use numerically approximated maxima.

Figures~\ref{FIG:TorusTube} and \ref{FIG:SphereTube} show the first
$n=4000$ replicate for the torus and sphere, respectively.  Each figure
restricts the fitted global tube to a geodesic, preserving its covariance
estimates and global cutoff.  The ellipses display how uncertainty changes
along the path.

\begin{figure}[!ht]
\centering
\includegraphics[width=0.92\textwidth]{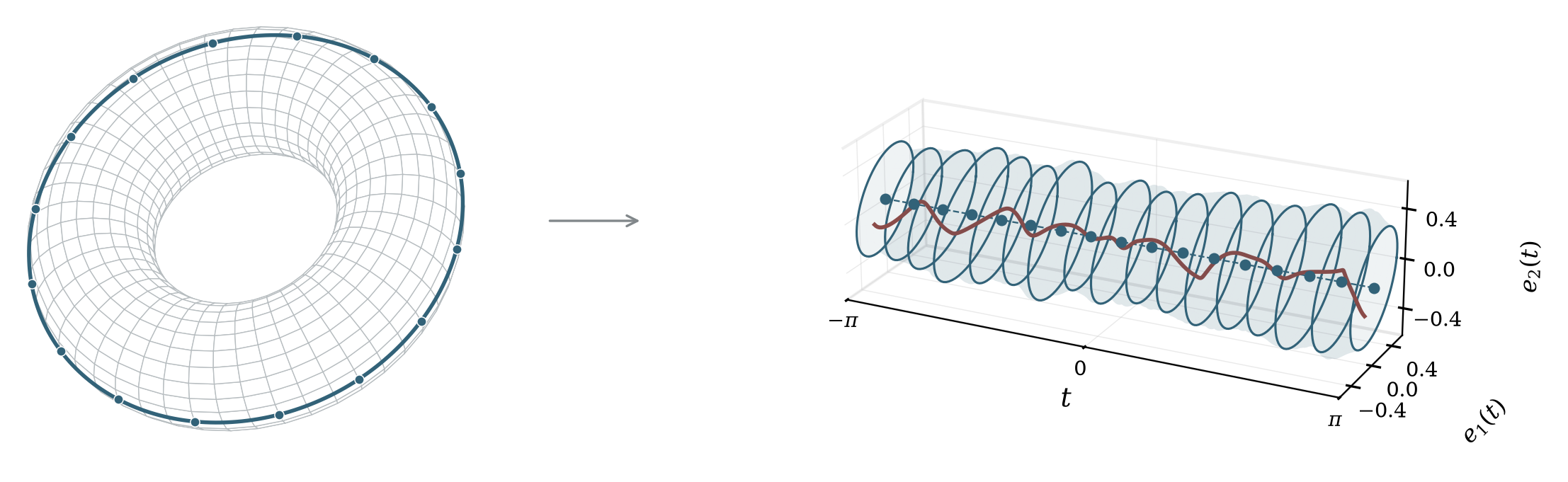}
\caption{A geodesic section of the nominal $95\%$ confidence tube on the flat
torus, using the first $n=4000$ production replicate.
Left: the blue path $\gamma\left(t\right)=\left(t,0\right)$ with the displayed evaluation points.
Right: matching blue ellipses bound the two-dimensional confidence regions
in $T_{\gamma\left(t\right)}\mathbb T^2$, translated to zero and arranged by $t$
in the orthonormal frame
$e_1\left(t\right)=\partial_{\theta_1}$, $e_2\left(t\right)=\partial_{\theta_2}$;
the red curve is $\widehat V_h\left\{\gamma\left(t\right)\right\}-V\left\{\gamma\left(t\right)\right\}$.
All cross-sections use one global analytic cutoff.
The endpoints $t=-\pi$ and $t=\pi$ coincide on the torus.
The embedding is schematic; estimation uses the flat product metric.}
\label{FIG:TorusTube}
\end{figure}

\begin{figure}[!ht]
\centering
\includegraphics[width=0.92\textwidth]{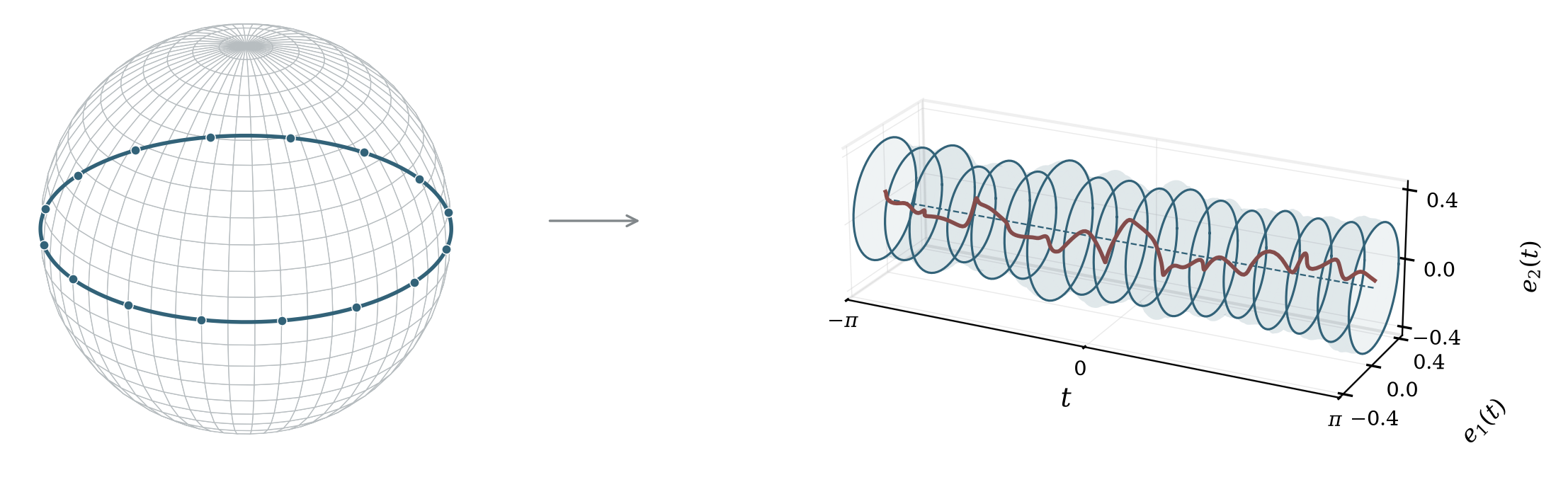}
\caption{An equatorial section of the nominal $95\%$ confidence
tube on $S^2$, using the first $n=4000$ production replicate.
Left: the blue equator $\gamma\left(t\right)=\left(\cos t,\sin t,0\right)$ and the displayed
locations.  Right: matching blue ellipses are zero-centered confidence
cross-sections in the parallel frame
$e_1\left(t\right)=\left(-\sin t,\cos t,0\right)$, $e_2\left(t\right)=\left(0,0,1\right)$;
the red curve is the estimation error in that frame.
Every ellipse uses the same global cutoff $4.700$.
The endpoints $t=-\pi$ and $t=\pi$ represent the same fibre.}
\label{FIG:SphereTube}
\end{figure}

\FloatBarrier
\endgroup
\section{Randomized reconstruction of planetary-scale 500-hPa wind}
\label{SEC:NCEPWind}
The simulations assess inference for known fields.  We now ask how accurately
a spatially smoothed mean wind can be reconstructed from randomized
space--month observations of a fixed climatological data set.  The uncertainty
comes from this randomized sampling, conditional on the observed wind fields.

\begin{figure}[!ht]
\centering
\includegraphics[width=\textwidth]{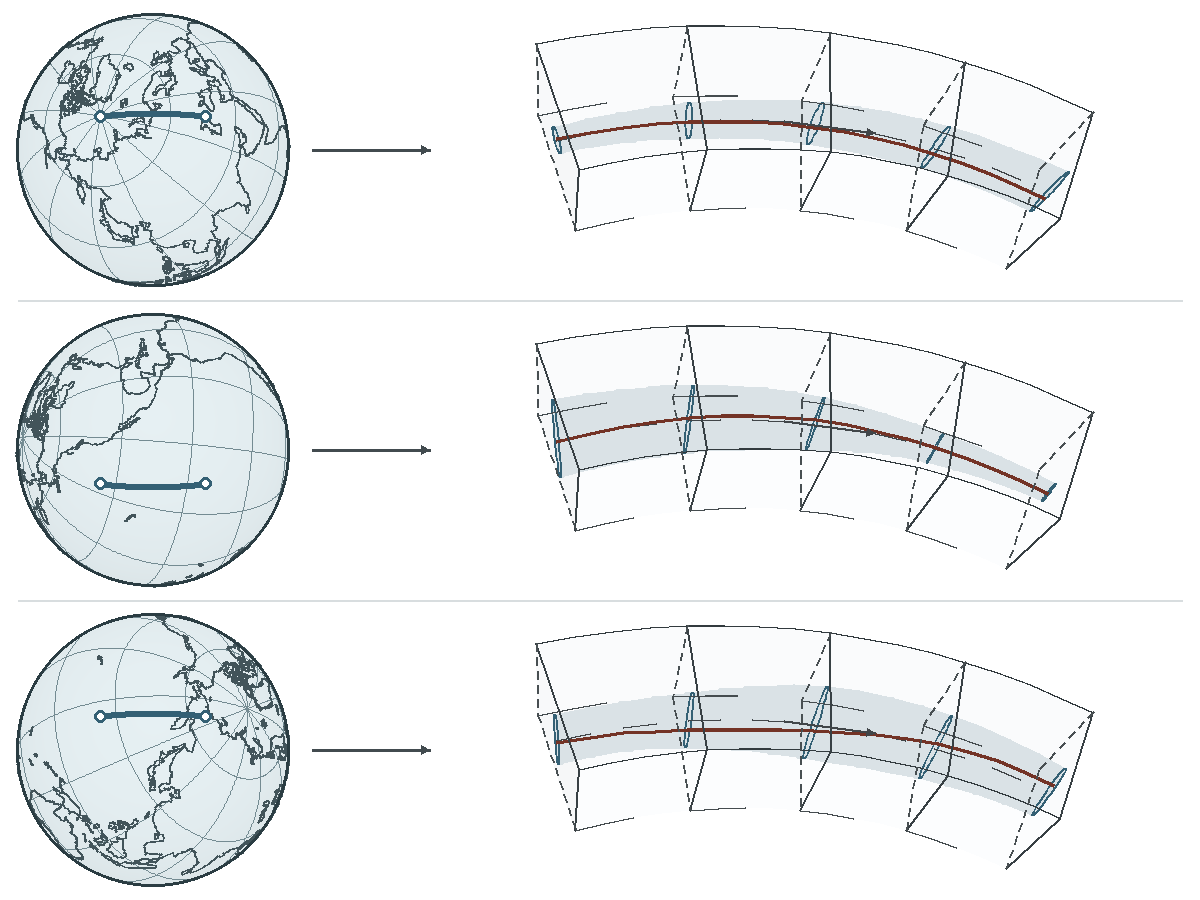}
\caption{Geodesic sections of the nominal $95\%$ analytic tube for the
smoothed mean.  Blue ellipses are zero-centered error sets and
the red curve is the realized reconstruction error.  The surface
interpolates 17 evaluated fibres along each path.  All rows share the
physical scale and the expanded cutoff $5.692$.}
\label{FIG:WindPathwiseTubes}
\end{figure}

We use the NCEP--NCAR Reanalysis~1 eastward and northward monthly mean winds
at 500 hPa for January 1991--December 2020, distributed by NOAA's Physical
Sciences Laboratory on a $2.5^\circ$ latitude--longitude grid
\citep{kalnay1996ncep}.  Each month's tangent field is projected by equal-area
weighted least squares onto the $16$ real electric and magnetic vector
spherical harmonics of degrees one and two.  Write $U_t\left(x\right)$ for the resulting
smooth field and
\[
 V_0\left(x\right)=\frac{1}{360}\sum_{t=1}^{360}U_t\left(x\right),\qquad
 \Sigma\left(x\right)=\frac{1}{360}\sum_{t=1}^{360}
                    \left\{U_t\left(x\right)-V_0\left(x\right)\right\}^{\otimes2}.
\]
Conditional on these fields, draw $n=4000$ independent uniform locations
$X_i\in S^2$ and independent uniform month indices $T_i\in\left\{1,\ldots,360\right\}$,
and observe $Y_i=U_{T_i}\left(X_i\right)$.  Then the pairs are iid,
$p=1/\left(4\pi\right)$, $\E\left(Y_i\mid X_i=x\right)=V_0\left(x\right)$, and the error covariance is
$\Sigma\left(x\right)$.  The responses retain units of metres per second.

Spatial resolution is part of the estimand.  Using the same volume-corrected
kernel as in the simulations, define
\begin{equation*}
 V_h\left(x\right)=\frac{\int_{S^2}W_h\left(x,q\right)\Tau_{q\to x}V_0\left(q\right)\dd\Vol\left(q\right)}{\int_{S^2}W_h\left(x,q\right)\dd\Vol\left(q\right)}.
\end{equation*}
The recorded choices are $h=c_hn^{-1/5}$ with $c_h=3,3.5,4$, corresponding
to $32.72^\circ$, $38.17^\circ$, and $43.63^\circ$ at $n=4000$.
The primary analysis uses $c_h=3$.  The other choices give smoother
targets by averaging over wider neighborhoods.  Each resolution is fixed
before calibration.

We use five folds assigned by observation index modulo five and fit each
training mean at the same bandwidth $h$ as the final mean.  The covariance
is centered at the target after transport:
\begin{equation}\label{EQ:TargetLocalStudentizer}
 \wh\Omega_{n,h}^{\rm TL}\left(x\right)
 =\frac{\sum_iW_h\left(x,X_i\right)^2
   \left\{\Tau_{X_i\to x}Y_i-\wh V_h^{\left(-k\left(i\right)\right)}\left(x\right)\right\}^{\otimes2}}{nh^2\left\{\wh p_h^\dagger\left(x\right)\right\}^2}.
\end{equation}
The target-local covariance in \eqref{EQ:TargetLocalStudentizer} includes
variation of the transported mean within the neighborhood as well as
measurement variation.  The supplementary material proves shrinking-bandwidth inference
for $V_h$ without undersmoothing.
It also gives a separate fixed-resolution Gaussian result with multiplier
calibration on a fixed finite grid.

The reconstruction uses the expanded analytic cutoff
$\beta_h^{\mathrm{vec}}+q_{0.95}/a_h=5.692375$ at the primary scale.
All displayed ellipsoid widths use this cutoff.  It is the expanded
counterpart of the unexpanded-centering cutoff in Algorithm~\ref{ALG:Tube}.
Evaluation on nested
icosahedral grids of $10242$ and $40962$ points gives a finest-grid maximum
studentized reconstruction error of $4.456$, below $5.692$.  The fifth percentile
of the effective-neighbor count is $85.8$, and the covariance floor is
inactive on the evaluation grid.  These diagnostics describe one randomized
reconstruction at the stated resolution.

Figure~\ref{FIG:WindPathwiseTubes} shows three geodesic sections of this
single global construction.  The cross-sections vary in size because the
estimated covariance varies over the sphere, although all use one cutoff.
Each path has length $45^\circ$.  From top to bottom, the endpoints are
the North Pole and $(45^\circ\mathrm E,45^\circ\mathrm N)$;
$(144.7^\circ\mathrm W,36.4^\circ\mathrm N)$ and
$(136.2^\circ\mathrm W,7.9^\circ\mathrm S)$; and
$(172.9^\circ\mathrm E,23.1^\circ\mathrm N)$ and
$(159.6^\circ\mathrm E,67.4^\circ\mathrm N)$.
The outer semi-axis changes from $0.49$ to $1.75\,\mathrm{m\,s^{-1}}$
on the first path, from $2.10$ to $0.56$ on the second, and from $1.43$
through a maximum of $2.07$ to $1.36$ on the third.

\FloatBarrier
\section*{Concluding remarks}

The proposed kernel estimator and confidence tube provide intrinsic
estimation and simultaneous inference for tangent vector fields.  Parallel
transport permits local averaging across tangent spaces, while a Gaussian
reference section supplies one critical value for the entire field.
Estimating the covariance locally allows the tube's orientation and width
to vary across the manifold.  The simulations show improving analytic
calibration with increasing sample size, and the wind reconstruction
illustrates the spatial variation in these uncertainty regions.

The theory assumes a known compact manifold without boundary, independent
observations and smooth uniformly nondegenerate covariance.  Small local
sample sizes can still limit the accuracy of the asymptotic calibration.
For chronological wind data, a natural next step is to replace the
independent-error covariance by a transported long-run covariance and establish
a Gaussian approximation under temporal dependence; the resulting spatial
correlation geometry must then be reanalyzed.  Bias correction could permit larger
regression neighborhoods, but would require uniform estimation of covariant
derivatives and a new covariance calculation for the corrected process.
\section*{Supplementary material}
Proofs and additional technical details are provided in the supplementary material.
\section*{Acknowledgment}
\par  Qirui Hu's research was supported by National Natural Science Foundation of China (NSFC) (Grant Nos.12601520), the Shanghai Engineering Research Center of Finance Intelligence (Grant No.~19DZ2254600) and by TRR 391 \textit{Spatio-temporal Statistics for the Transition of Energy and Transport} (Project number 520388526) funded by the Deutsche Forschungsgemeinschaft (DFG, German Research Foundation).
\bibliographystyle{plainnat}
\bibliography{references}

\end{document}